%% file: aaai2027.tex
\documentclass[letterpaper]{article} 
\usepackage[preprint]{aaai2027}  
\usepackage[hyphens]{url} 
\usepackage{graphicx}
\usepackage{natbib}
\usepackage{caption}
\usepackage{booktabs}
\input{package}

\usepackage{newfloat}
\usepackage{listings}
\usepackage{tikz}

\usepackage{array}
\usepackage[table]{xcolor}
\usepackage{tabularx}

\definecolor{rankfirst}{RGB}{0,120,80}
\definecolor{ranksecond}{RGB}{30,95,160}
\definecolor{rankthird}{RGB}{200,95,35}

\definecolor{corecolbg}{gray}{0.975}
\definecolor{corefirstbg}{RGB}{221,240,230}
\definecolor{coresecondbg}{RGB}{226,236,248}
\definecolor{corethirdbg}{RGB}{249,234,222}

\DeclareCaptionStyle{ruled}{labelfont=normalfont,labelsep=colon,strut=off} 
\floatstyle{ruled}
\newfloat{listing}{tb}{lst}{}
\floatname{listing}{Listing}
\title{Towards Anomaly Detection on Relational Data}

\author{
    Shiyuan Li\textsuperscript{\rm 1},
    Yunfeng Zhao\textsuperscript{\rm 2},
    Yue Tan\textsuperscript{\rm 1},
    Qingfeng Chen\textsuperscript{\rm 2},
    Yixin Liu\textsuperscript{\rm 1},
    Shirui Pan\textsuperscript{\rm 1}\corresponding
}
\affiliations{
    \textsuperscript{\rm 1}Griffith University\\
    \textsuperscript{\rm 2}Guangxi University\\
    \{shiyuan.li, yue.tan,  yixin.liu, s.pan\}@griffith.edu.au,\\
    \{yunf.zhao, qingfeng\}@gxu.edu.cn

}

\begin{document}

\maketitle

\begin{abstract}
\input{Sections/0_abs}
\end{abstract}

\section{Introduction}
\input{Sections/1_intro}

\section{Related Work}
\input{Sections/2_rw}

\section{Preliminaries}
\input{Sections/3_preliminary}

\section{Methodology}
\input{Sections/4_method}

\section{Experiments}
\input{Sections/5_exp}

\section{Conclusion}
\input{Sections/6_con}

\bibliography{ref}


\clearpage

\appendix

\setcounter{secnumdepth}{2}
\setcounter{section}{0}
\renewcommand{\thesection}{\Alph{section}}
\renewcommand{\thesubsection}{\Alph{section}.\arabic{subsection}}
\renewcommand{\thesubsubsection}{\Alph{section}.\arabic{subsection}.\arabic{subsubsection}}
\input{Sections/7_appendix}

\end{document}

%% file: package.tex
\usepackage[utf8]{inputenc}
\usepackage[T1]{fontenc}
\usepackage{float}
\usepackage{url}
\usepackage{booktabs}
\usepackage{amsmath}
\usepackage{amssymb}
\usepackage{amsfonts}
\usepackage{graphicx}
\usepackage{xcolor}
\usepackage{microtype}
\usepackage{nicefrac}
\usepackage{xspace}
\usepackage{multirow}
\usepackage{subfigure}
\usepackage{pifont}
\usepackage{enumitem}
\usepackage{color, colortbl}

\definecolor{Gray}{gray}{0.9}
\definecolor{fgreen}{RGB}{177,207,149}
\definecolor{fred}{RGB}{234,179,138}
\definecolor{firstcolor}{RGB}{20,128,85}
\definecolor{secondcolor}{RGB}{20,104,168}
\definecolor{thirdcolor}{RGB}{236,84,20}

\newcommand{\ourmethod}{{RelAD}\xspace}

\usepackage[linesnumbered,algoruled,boxed,lined,ruled]{algorithm2e}
\SetKwInOut{Param}{Parameters}

\SetCommentSty{mycommfont}

%% file: Sections/0_abs.tex
Relational databases are widely used for managing structured data in real-world systems. Detecting anomalies from such relational data is crucial for identifying fraud, risks, and abnormal behaviors, yet remains under-explored. The key challenges lie in the intrinsic complexity of relational data: multi-table attributes are high-dimensional and heterogeneous, making sparse abnormal clues easy to overwhelm by normal or irrelevant information. Moreover, anomalies may further manifest as abnormal connection patterns across different foreign-key relations, which existing tabular and graph anomaly detection methods are ill-suited to capture. To address them, we propose \ourmethod, a reconstruction-based framework that captures anomalies from both attribute and relational edge reconstruction. \ourmethod contains two core modules: conditional sparse-gated attribute reconstruction, which suppresses redundant multi-table attributes and emphasizes abnormal semantic blocks, and dual-view multi-relational edge reconstruction, which detects relation-specific abnormal connections from both intrinsic and behavioral instance profiles. The resulting attribute and relational signals are integrated through a lightweight fusion module to produce the final anomaly score. We further construct 6 benchmark datasets with systematic anomalies, on which extensive experiments show that \ourmethod consistently outperforms baselines while achieving competitive efficiency.

%% file: Sections/1_intro.tex
Relational databases are widely used as the primary storage for structured data in real-world applications such as financial risk control, e-commerce, and industrial operation monitoring~\cite{robinson2024relbench,dwivedi2025relational}. Unlike single-table data, a relational database organizes information across multiple tables interconnected via primary--foreign key relationships, modeling rich structural and temporal dependencies between samples. For example, a user in a risk-control system may link to products, devices, and transactions across multiple tables~\cite{dwivedi2025relationalGT,chen2025relgnn}. 
To enable automatic knowledge discovery from relational databases, recent studies on relational deep learning (RDL) develop data-driven neural models to capture cross-table dependencies, demonstrating strong capability for representation learning on relational databases~\cite{robinson2024relbench,ranjan2025relational,wang2025griffin}. 

\begin{figure}[!t]
\centering
\includegraphics[width=\columnwidth]{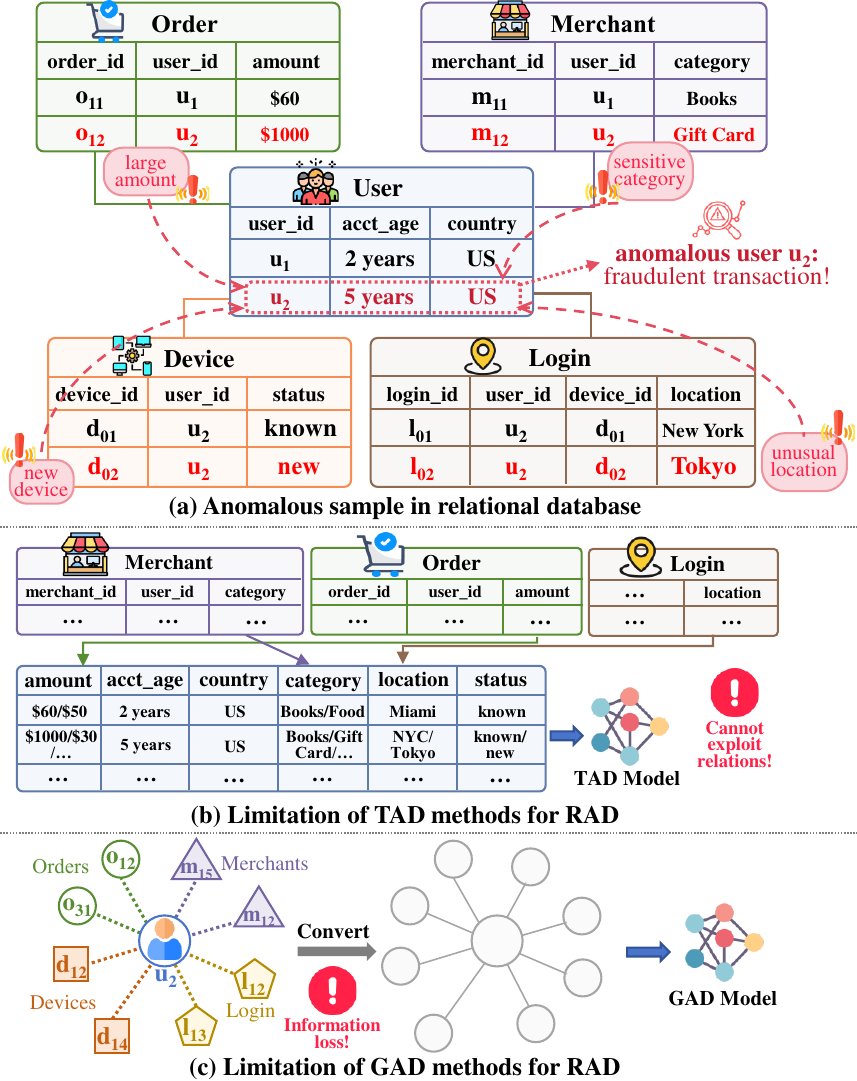}
\caption{Sketch maps of anomalies in relational databases and the limitations of current methods.}
\label{fig:intro_motivation}
\end{figure}

With the widespread use of relational databases, various anomalies also arise from data entry errors, system failures, or unexpected real-world events~\cite{dwivedi2025relational}. 
For example, as shown in Fig.~\ref{fig:intro_motivation}(a), fraudulent transactions may only become identifiable through the joint examination of records across multiple tables, such as users, devices, accounts, and products, while each record may appear normal when viewed in isolation~\cite{li2026towards,pan2025label}.
To ensure data reliability and support trustworthy downstream decision-making, it is crucial to identify such anomalies from relational databases. Despite its practical importance, \textit{relational anomaly detection (RAD)} remains largely under-explored and has received limited attention from existing anomaly detection and RDL communities.

To address this under-explored problem, 
a straightforward solution is to transform the database into a format compatible with existing anomaly detection methods, such as \textbf{tabular anomaly detection} (TAD)~\cite{thimonier2024beyond,ye2025drl} or \textbf{graph anomaly detection} (GAD)~\cite{pan2026correcting,pan2025survey,ding2019deep,liu2021anomaly,chen2024boosting,pan2026camera}, and directly adapt them to RAD. 
Specifically, in order to apply TAD to relational databases, we can flatten relational schemas into single tables through feature engineering and aggregation (as in Fig.~\ref{fig:intro_motivation}b).
Although it is straightforward, the flattening will discard the dependencies encoded by primary--foreign key relationships, leading to information loss. Meanwhile, the resulting high-dimensional, redundant feature matrix can dilute local abnormal signals in global objectives. 
On the other hand, while GAD naturally captures inter-sample connections in relational data, most existing GAD methods are designed for homogeneous graphs~\cite{liu2024arc,zhao2025freegad}, and thus struggle to distinguish the diverse relation types induced by different primary--foreign key relationships. 
Directly converting a relational database into a homogeneous graph (as in Fig.~\ref{fig:intro_motivation}c) would mix relation-specific semantics, making it difficult to identify which relational patterns lead to the abnormality.

These limitations motivate a dedicated framework for RAD, which faces two unique challenges.
\textit{\textbf{C1}~-~Feature redundancy and signal dilution.} Unlike standard tabular data with carefully curated features from a single table, relational databases are inherently multi-table and heterogeneous: a central instance can be described by attributes in the central table, attributes from child tables, and feature groups derived from different instance contexts. This naturally leads to a high-dimensional and redundant feature space in which only a small fraction of attributes can indicate abnormal behavior. Consequently, the massive volume of anomaly-irrelevant features tends to dominate the model's learning objective, diluting or completely obscuring subtle local abnormal signals. In this case, how to adaptively extract sparse anomaly indicators from overwhelming redundant features poses a significant challenge. \textit{\textbf{C2}~-~Relational heterogeneity and complex cross-table dependencies.} Anomalies in relational databases are often embedded in complex dependencies, where a central instance interacts with different types of neighbor instances through semantically distinct primary--foreign key relationships. More importantly, different types of relations may contribute unequally to the abnormality of an instance, requiring the model to selectively focus on the most informative dependencies. In this context, exploiting heterogeneous relations to capture relation-specific abnormal connection patterns requires effective modeling of relation semantics. 
In this context, how to exploit heterogeneous relations to capture relation-specific abnormal connection patterns poses another key challenge.

To tackle these challenges, we propose \textbf{\underline{Rel}}ational \textbf{\underline{A}}nomaly \textbf{\underline{D}}etection (\ourmethod), a reconstruction-based framework for RAD. \ourmethod is specifically designed for relational databases, with the core idea of characterizing anomalies from both attribute reconstruction and relational edge reconstruction. To address \textit{\textbf{C1}}, \ourmethod incorporates a conditional sparse-gated attribute reconstruction module. This module generates conditional masks for central-table attributes and attributes aggregated from child tables according to the block structure of relational databases, enabling the model to adaptively select informative dimensions before reconstruction. It further adopts block-specific decoding and block-level residuals, allowing anomaly scoring to focus on the most significant local attribute deviations instead of being diluted by global average errors. To address \textit{\textbf{C2}}, \ourmethod proposes a dual-view multi-relational edge reconstruction module, which reconstructs relation-specific edges over the heterogeneous graph induced by primary--foreign key relationships. Specifically, the model encodes each central instance using both its central-table profile and an aggregated profile derived from the associated child tables.
It learns relation-specific neighbor-instance representations for each relation type. 
In this way, \ourmethod measures whether an instance profile can explain its connection behaviors under different relations. Finally, \ourmethod fuses three complementary signals, including attribute-block anomalies, self-profile relational anomalies, and child-profile relational anomalies, to produce the final anomaly score and cover diverse anomaly sources. 
In summary, this paper makes the following contributions:

\begin{itemize}[noitemsep,leftmargin=*,topsep=1.5pt]
    \item \textbf{Problem.} 
    To the best of our knowledge, we are the first to formalize the problem of relational anomaly detection. For evaluation, we design domain-specific anomaly synthesis rules for each dataset to build a comprehensive benchmark.
    \item \textbf{Methodology.} 
    We propose \ourmethod, a novel reconstruction-based method that jointly captures local attribute deviations and relation-specific abnormal connection patterns through attribute and multi-relational edge reconstruction.
    \item \textbf{Experiments.} 
    Extensive experiments on six benchmark datasets validate the effectiveness, robustness, and efficiency of \ourmethod.
\end{itemize}

%% file: Sections/2_rw.tex
In this section, we briefly review three lines of related studies. A detailed literature review is in Appendix~\ref{app:rw}.

\noindent\textbf{Relational Deep Learning (RDL)} aims to learn from multi-table relational databases by modeling instances and their primary--foreign key relationships~\cite{robinson2024relational}. Mainstream studies transform relational databases into heterogeneous graphs and develop graph neural network (GNN)-based or Transformer-based frameworks for representation learning and downstream task modeling~\cite{dwivedi2025relationalGT,chen2025relgnn}. Advanced studies have also extended RDL to more general paradigms, such as LLM-based frameworks~\cite{wu2025large} and foundation models~\cite{ranjan2025relational,wang2025griffin,kothapalli2026plurel} for RDL. Despite their success, existing works rarely investigate anomaly detection for relational data, leaving this problem largely under-explored.

\noindent\textbf{Tabular Anomaly Detection (TAD)} investigates the detection of anomalous instances in structured tabular data~\cite{pang2021deep, borisov2022deep}. Early TAD methods rely on handcrafted statistical assumptions, including isolation-, density-, and distance-based anomaly scoring strategies~\cite{breunig2000lof,ramaswamy2000efficient,liu2019generative}. Recent deep learning-based solutions, built upon techniques such as autoencoding and contrastive learning, demonstrate strong performance in TAD~\cite{ye2025drl,yin2024mcm,shenkar2022anomaly}. However, TAD methods cannot effectively model the relational dependencies in relational data, making them sub-optimal for RAD. 

\noindent\textbf{Graph Anomaly Detection (GAD)} aims to identify anomalous samples from graph data~\cite{ma2021comprehensive,qiao2025deep,liu2026few,shen2026raising,zhao2026fedcigar,liu2026rethinking}. As shallow methods usually struggle to handle complex graph data~\cite{peng2018anomalous,li2017radar}, mainstream studies primarily develop deep learning-based approaches built upon GNNs~\cite{qiao2025deep}. These deep methods mainly rely on paradigms such as reconstruction~\cite{ding2019deep}, contrastive learning~\cite{liu2021anomaly,pan2023prem}, and affinity learning~\cite{qiao2023truncated} to characterize anomalous patterns. Nevertheless, most GAD methods are developed for homogeneous graphs, limiting their effectiveness in RAD with heterogeneous instances and relations.

%% file: Sections/3_preliminary.tex
\noindent\textbf{Relational Database.}
Following relational deep learning, we consider a relational database as a collection of tables $\mathcal{D}=\{\mathcal{T}^{k}\}_{k=1}^{K}$, where $K$ is the number of tables. Each table $\mathcal{T}^{k}$ contains rows corresponding to instances and columns for their attributes. Tables are connected by primary--foreign key relationships. A primary key uniquely identifies a row in a table, while a foreign key in another table references this primary key and thereby defines a typed relationship between rows from two tables. Thus, relational databases contain both table attributes and inter-table instance relationships.

\noindent\textbf{Heterogeneous Graph View.}
A relational database can be represented as a heterogeneous graph $\mathcal{G}=(\{\mathcal{V}^{k}\}_{k=1}^{K},\{\mathcal{E}_{r}\}_{r\in\mathcal{R}})$, where each node set $\mathcal{V}^{k}$ is induced by the rows of table $\mathcal{T}^{k}$. The set $\mathcal{R}$ contains relation types derived from the database schema. Specifically, each relation type $r\in\mathcal{R}$ corresponds to a primary--foreign key constraint, or equivalently a typed link between a source table, a foreign-key column, and a referenced table. The associated edge set $\mathcal{E}_{r}$ is obtained by instantiating this constraint over the rows of the two tables. An edge $(v,i)\in\mathcal{E}_{r}$ indicates that two rows are connected under relation type $r$. Different relation types preserve different semantics, such as user--item interactions, user--device associations, or paper--citation links. This heterogeneous graph view is only used to define the relational structure; our setting keeps relation-specific edge sets instead of collapsing them into a single homogeneous graph.

\noindent\textbf{Relational Anomaly Detection Setting.}
For RAD, we designate one table as the target, or central, table whose instances are to be detected, and refer to schema-connected tables as child or related tables when describing instance-centric features and neighborhoods. We focus on identifying abnormal instances from the central table $\mathcal{T}^{0}$. Let $\mathcal{U}=\{u_1,\dots,u_N\}$ denote the $N$ central instances. Each central instance $u\in\mathcal{U}$ may be connected to heterogeneous neighbor instances through multiple relation types. We represent the relation-specific edge set incident to central instances as:
\begin{equation}
\label{eq:edge_set}
\mathcal{E}=\{(u,r,i)\mid u\in\mathcal{U},\; r\in\mathcal{R},\; i\in\mathcal{V}^{r}\},
\end{equation}
\noindent where $\mathcal{V}^{r}$ denotes the neighbor-instance table associated with relation type $r$, and $(u,i)\in\mathcal{E}_{r}$ indicates that $u$ is connected to neighbor instace $i$ under relation $r$. An anomaly label vector $\mathbf{y}\in\{0,1\}^{N}$ is used for evaluation, where $\mathbf{y}_u=1$ indicates that target instance $u$ is anomalous and $\mathbf{y}_u=0$ otherwise. The goal is to learn an anomaly scoring function $f:\mathcal{U}\rightarrow\mathbb{R}$ such that anomalous instances receive larger scores than normal instances. At a high level, anomalies may appear as abnormal attribute patterns of a target instance, abnormal relation-specific connection patterns, or inconsistencies between instance attributes and relational connections.

\noindent\textbf{Initial Features.}
For each table, raw columns can include numerical, categorical, timestamp, and text attributes. Following the feature preprocessing protocol of RelBench~\cite{gu2026relbench}, we encode heterogeneous raw columns into unified row-level representations, where non-text attributes are normalized or embedded according to their types and text attributes are represented by pretrained text embeddings.  
For each target instance $u$, we construct an attribute vector:
\begin{equation}
\label{eq:x_attr}
\mathbf{x}_{u}=\operatorname{Concat}\left(\mathbf{x}_{u}^{self},\mathbf{x}_{u}^{agg,1},\dots,\mathbf{x}_{u}^{agg,B}\right)\in\mathbb{R}^{d},
\end{equation}
\noindent where $\mathbf{x}_{u}^{self}$ is the row feature of $u$ from the target table, and each $\mathbf{x}_{u}^{agg,b}$ is an aggregated feature block computed from rows in related tables. Here, $B$ denotes the number of child-table aggregated blocks, so the complete attribute vector contains $B+1$ blocks including the self block. The aggregation uses statistics such as mean, standard deviation, and count over rows connected to $u$. The resulting matrix is denoted by $\mathbf{X}\in\mathbb{R}^{N\times d}$. For relation-specific edge reconstruction, each relation type $r$ additionally provides an edge list $\mathcal{E}_{r}$ and a neighbor feature matrix $\mathbf{H}^{r}\in\mathbb{R}^{|\mathcal{V}^{r}|\times d_r}$, where $d_r$ is the feature dimension of neighbor instances under relation $r$.

%% file: Sections/4_method.tex
\begin{figure*}[!t]
\centering
\includegraphics[width=\textwidth]{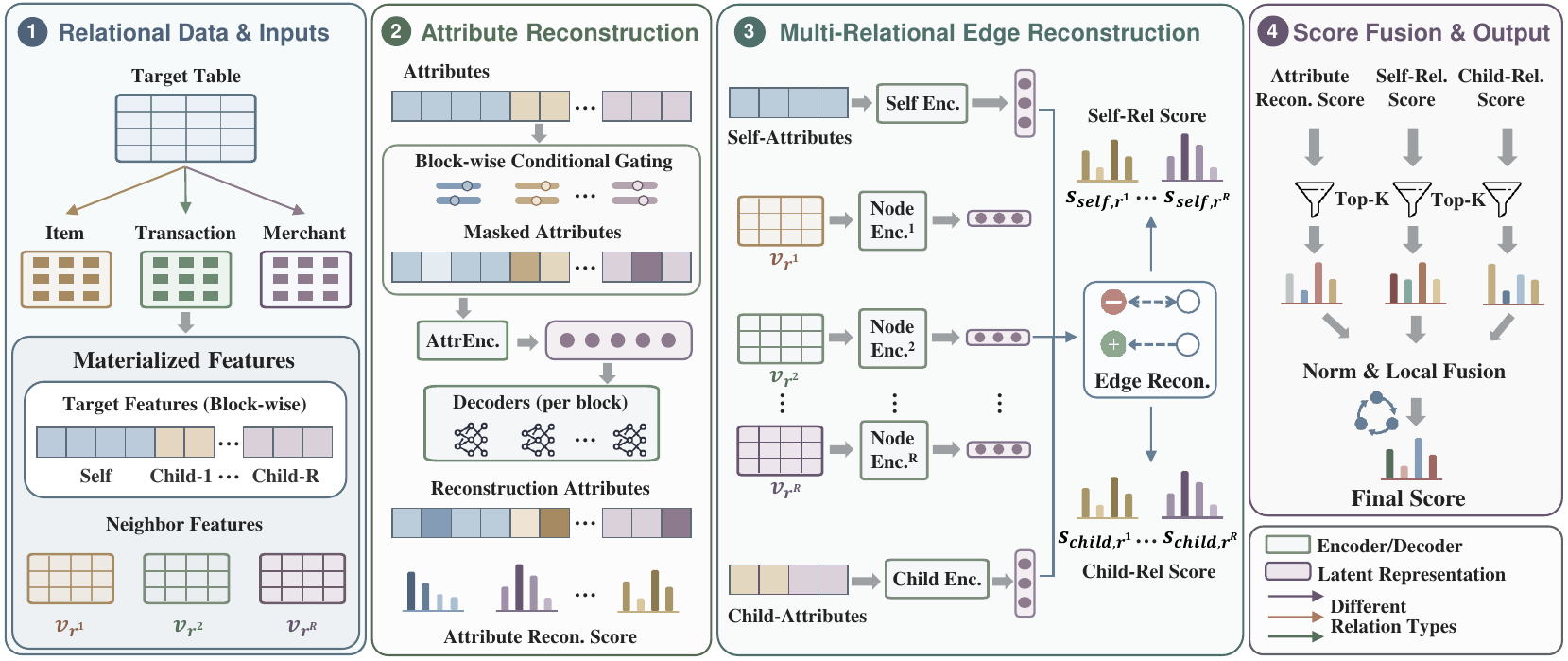}
  \caption{The overall pipeline of \ourmethod for relational data anomaly detection. 
  }
  \label{fig:pipeline}
\end{figure*}

In this section, we introduce \ourmethod, a reconstruction-based framework for anomaly detection on relational data. 
Since anomaly labels are rarely available and normal instances dominate the data, an unsupervised reconstruction model can learn normal patterns from unlabeled instances, while instances deviating from these patterns incur larger reconstruction residuals, making reconstruction error a natural anomaly indicator.
Based on this principle, \ourmethod characterizes anomalies from two complementary views, as shown in Fig.~\ref{fig:pipeline}. To suppress redundant cross-table attributes and preserve local semantic deviations, we first introduce a \textit{conditional sparse-gated attribute reconstruction} module. Next, to model relation-specific abnormal connections under foreign keys, we propose a \textit{dual-view multi-relational edge reconstruction} module, which reconstructs relational edges from both the intrinsic instance profile and the child-table behavioral profile. Finally, we combine the most salient attribute and relational signals through an \textit{anomaly score fusion} strategy.

\subsection{Conditional Sparse-Gated Attribute Reconstruction}\label{subsec:cgar}

In relational databases, a central instance can be represented using heterogeneous attributes from its central table and multiple child tables. These attributes naturally form semantic blocks with different relevance to anomaly detection. Since abnormal signals may only appear in a few blocks or dimensions, directly reconstructing the entire attribute vector can make numerous anomaly-irrelevant features dominate the objective. To address this challenge, \ourmethod introduces a conditional sparse-gated attribute reconstruction module, which first identifies informative relational feature blocks and then reconstructs the instance profile.

\noindent\textbf{Conditional Sparse Gating.}
To suppress anomaly-irrelevant attributes before reconstruction, a gating mechanism can assign adaptive importance to input dimensions. However, a single global gate treats all dimensions as an unstructured feature set and ignores block semantics induced by different tables. Moreover, whether a child-table block is informative often depends on the central instance profile rather than the child block alone, requiring explicit cross-block conditioning. To this end, we design a block-aware and context-conditioned gate that generates a separate mask per semantic block. Let $\mathbf{x}^{(0)}$ be the central-table block and $\mathbf{x}^{(t)}$ the $t$-th child-table aggregated block. The central block mask is generated from its own attributes, while each child block mask is conditioned on the central block to determine whether the related-table information is informative:
\begin{equation}
\label{eq:mask_child}
\mathbf{m}^{(t)} = \begin{cases}
\sigma(\operatorname{MLP}_0(\mathbf{x}^{(0)})), & t=0 \\
\sigma(\operatorname{MLP}_t([\mathbf{x}^{(0)};\mathbf{x}^{(t)}])), & t=1,\dots,B
\end{cases}
\end{equation}
\noindent where $B$ is the number of child-table aggregated blocks, $\mathbf{m}^{(0)}$ and $\mathbf{m}^{(t)}$ are block-level masks, and $\sigma(\cdot)$ is the sigmoid function. After obtaining these block-level masks, we concatenate them to form an instance-specific attribute mask aligned with the original feature vector for input feature filtering:
\begin{equation}
\label{eq:gated_input}
\mathbf{m}=[\mathbf{m}^{(0)};\mathbf{m}^{(1)};\dots;\mathbf{m}^{(B)}], \qquad \tilde{\mathbf{x}}=\mathbf{m}\odot\mathbf{x},
\end{equation}
\noindent where $\mathbf{m}$ has the same dimensionality as $\mathbf{x}$, $\tilde{\mathbf{x}}$ is the gated attribute vector, and $\odot$ denotes element-wise multiplication. Thus, the encoder receives a compact attribute profile that emphasizes informative blocks and suppresses redundancy. The gated vector is then projected into a latent representation: $\mathbf{z}_{attr} = \operatorname{Enc}_{attr}(\tilde{\mathbf{x}})$, where $\mathbf{z}_{attr}$ is the attribute latent representation and $\operatorname{Enc}_{attr}(\cdot)$ is the shared attribute encoder.

Although conditional gating assigns adaptive importance to different attributes, the learned masks may remain dense without explicit regularization. To encourage sparse attribute selection, we impose a sparsity penalty on mask values:
\begin{equation}
\label{eq:sparse_loss}
\mathcal{L}_{sparse} = \frac{1}{Nd}\sum_{u=1}^{N}\sum_{j=1}^{d}\mathbf{m}_{u,j},
\end{equation}
\noindent where $\mathbf{m}_{u,j}$ denotes the gate value of instance $u$ on the $j$-th attribute dimension. With this constraint, the model is encouraged to reconstruct instance profiles using a compact subset of informative attributes, reducing the interference from redundant dimensions.

\noindent\textbf{Block-Specific Reconstruction.}
After obtaining $\mathbf{z}_{attr}$, a straightforward reconstruction strategy is to use a single decoder to predict the entire attribute vector. However, such a decoder ignores that different blocks originate from different tables and carry distinct semantics. It may also allow high-dimensional blocks to dominate the reconstruction objective, making anomalies in small but important blocks less visible.  To preserve block-level semantics during reconstruction, \ourmethod adopts block-specific decoders, where each decoder is responsible for reconstructing one semantic block:
\begin{equation}
\label{eq:block_decoder}
\hat{\mathbf{x}}^{(t)} = \operatorname{Dec}_t(\mathbf{z}_{attr}), \quad t=0,\dots,B,
\end{equation}
\noindent where $\hat{\mathbf{x}}^{(t)}$ is the reconstructed vector of block $t$ and $\operatorname{Dec}_t(\cdot)$ is the decoder for that block. The reconstructed blocks are then concatenated to form the full reconstruction $\hat{\mathbf{x}}_u$ for each instance. During training, we optimize the overall reconstruction objective over the full attribute vector:
\begin{equation}
\label{eq:attr_rec_loss}
\mathcal{L}_{rec} = \frac{1}{Nd}\sum_{u=1}^{N}\left\|\mathbf{x}_u-\hat{\mathbf{x}}_u\right\|_2^2,
\end{equation}
\noindent where $\hat{\mathbf{x}}_u$ is the reconstructed full attribute vector of instance $u$. While this objective trains the attribute branch to reconstruct the complete instance profile, anomaly scoring should remain sensitive to localized deviations. Therefore, for inference, we further compute a block-level reconstruction error:
\begin{equation}
\label{eq:block_error}
\mathcal{L}_t(u) = \frac{1}{d_t}\left\|\mathbf{x}_{u}^{(t)}-\hat{\mathbf{x}}_{u}^{(t)}\right\|_2^2,
\end{equation}
\noindent where $t=0,\dots,B$, $\mathbf{x}_{u}^{(t)}$ and $\hat{\mathbf{x}}_{u}^{(t)}$ are the original and reconstructed vectors of block $t$, and $d_t$ is the dimensionality of this block. These block-wise residuals reveal which semantic parts of an instance are poorly reconstructed. Since anomalies may only appear in a few semantic blocks, aggregating all block errors uniformly may dilute the abnormal signal. We therefore use the average of the top-$K$ normalized block errors as the local attribute anomaly score:
\begin{equation}
\label{eq:attr_topk}
s_{attr}(u) = \frac{1}{K}\sum_{t \in \operatorname{TopK}_{t}(\operatorname{Norm}(\mathcal{L}_t(u)))}\operatorname{Norm}(\mathcal{L}_t(u)),
\end{equation}
\noindent where $\operatorname{Norm}(\cdot)$ denotes score normalization, $K$ is a hyperparameter controlling the number of selected local signals, and $\operatorname{TopK}_{t}(\cdot)$ selects the $K$ largest block-level scores. As a result, the attribute branch produces an anomaly score that is less affected by redundant or weakly relevant attributes and more sensitive to localized deviations within specific semantic blocks. This allows \ourmethod to distinguish instances with abnormal attribute patterns even when such deviations are sparse and would be diluted by global reconstruction errors or uniform block aggregation.

\subsection{Dual-View Multi-Relational Edge Reconstruction}\label{subsec:dmer}

Apart from attributes, relations also play a critical role in RAD, as abnormal instances may form abnormal connections across multiple foreign-key relations, each revealing a different type of behavioral deviation.
The abnormality can arise from two sources: an edge inconsistent with the instance's intrinsic central-table profile, or one inconsistent with the behavioral context from child tables.
Collapsing these into a single representation may blur their distinct roles, as behavioral aggregations can provide shortcut signals that suppress self-profile evidence. \ourmethod therefore reconstructs multi-relational edges from both views via self-profile and child-profile branches.

\noindent\textbf{Dual-View Instance Encoding.}
To keep the two sources of relational evidence disentangled, we encode the central-table profile and the child-table aggregations with two independent encoders. The self-profile branch maps the central-table profile into a latent space, allowing the model to examine whether the observed neighbors are compatible with the intrinsic properties of the target instance:
\begin{equation}
\label{eq:rel_self_encoder}
\mathbf{z}_{u,self}^{rel} = \operatorname{Enc}_{self}(\mathbf{x}_{u}^{self}),
\end{equation}
\noindent where $\mathbf{z}_{u,self}^{rel}$ is the relational representation from the central-table profile and $\operatorname{Enc}_{self}(\cdot)$ is the corresponding encoder. In parallel, the child-profile branch encodes the aggregated child-table profile, allowing edge reconstruction to be conditioned on the target instance's historical behavioral context:
\begin{equation}
\label{eq:rel_child_encoder}
\mathbf{z}_{u,child}^{rel} = \operatorname{Enc}_{child}(\mathbf{x}_{u}^{child}),
\end{equation}
\noindent where $\mathbf{z}_{u,child}^{rel}$ is the relational representation from child-table aggregations and $\operatorname{Enc}_{child}(\cdot)$ is an independent encoder. By keeping the two encoders separate, \ourmethod can produce view-specific relational residuals rather than collapsing all evidence into a single mixed representation.

After obtaining the two target-instance views, the neighbor side should still preserve relation semantics, because different foreign-key relations may connect the target instance to different types of neighbor records. Therefore, for each relation type $r \in \mathcal{R}$, \ourmethod uses a relation-specific neighbor-instance encoder shared by both branches:
\begin{equation}
\label{eq:neighbor_entity_encoders}
\mathbf{e}_i^{r} = \operatorname{NodeEnc}_{r}(\mathbf{h}_i^{r}),
\end{equation}
\noindent where $\mathbf{h}_i^{r}$ is the row feature of neighbor instance $i$ in $\mathbf{H}^{r}$, $\operatorname{NodeEnc}_{r}(\cdot)$ is the neighbor-instance encoder of relation $r$, and $\mathbf{e}_i^{r}$ is the shared relation-specific neighbor-instance embedding used by both branches. Sharing this neighbor embedding space makes the reconstruction scores from the two views comparable within each relation, while the relation-specific encoder prevents semantically different foreign-key relations from being forced into a single neighbor representation.

\noindent\textbf{Multi-Relational Edge Reconstruction.}
Given an observed edge $(u,i)\in\mathcal{E}_r$ and a randomly sampled negative neighbor instance $j$, the two branches reconstruct relational edges via dot-product scoring. Let $q\in\{self,child\}$ denote the relational branch, with $\mathbf{z}_{u,q}^{rel}$ being the corresponding target-instance representation. The branch-specific relational loss can be written as:
\begin{equation}
\label{eq:rel_branch_loss}
\begin{split}
\mathcal{L}_{q}^{r} = \;& \mathbb{E}_{(u,i)\in\mathcal{E}_r}\left[\phi(-\mathbf{z}_{u,q}^{rel}\cdot \mathbf{e}_i^{r})\right] \\
& + \mathbb{E}_{(u,j)\notin\mathcal{E}_r}\left[\phi(\mathbf{z}_{u,q}^{rel}\cdot \mathbf{e}_j^{r})\right],
\end{split}
\end{equation}
\noindent where $\phi(x)=\log(1+\exp(x))$ denotes the softplus function.

To obtain anomaly scores during inference, for each relation type, we compute the average edge-level negative log-likelihood for each instance and normalize the relation-wise scores. The top-$K$ relation scores are then aggregated for each branch:
\begin{equation}
\label{eq:rel_topk}
s_{rel,q}(u)=\frac{1}{K}
\sum_{r\in\mathcal{R}_{K,q}(u)}
\operatorname{Norm}(\bar{\ell}_{q}^{r}(u)),
\quad q\in\{self,child\},
\end{equation}
\noindent where $\bar{\ell}_{q}^{r}(u)$ denotes the average positive-edge negative log-likelihood under relation type $r$ from branch $q$, and $\mathcal{R}_{K,q}(u)=\operatorname{TopK}_{r}\{\operatorname{Norm}(\bar{\ell}_{q}^{r}(u))\}$ selects the top-$K$ relations by normalized score. Equipped with dual-view reconstruction, \ourmethod captures inconsistencies between relational behaviors and both central-profile and child-table contexts.

\input{Tables/main_result}

\subsection{Model Training and Anomaly Scoring}\label{subsec:train_eval}

To optimize \ourmethod, the overall objective combines the previous loss terms:
\begin{equation}
\label{eq:overall_loss}
\mathcal{L} = \mathcal{L}_{rec} + \lambda_s \mathcal{L}_{sparse} + \frac{1}{|\mathcal{R}_{act}|}\sum_{q\in\{self,child\}}\sum_{r\in\mathcal{R}_{act}}\mathcal{L}_{q}^{r},
\end{equation}
\noindent where $\mathcal{R}_{act}$ is the set of active relation types with valid sampled edges in the current batch, and $\lambda_s$ controls the sparsity regularization. 

\noindent\textbf{Instance Anomaly Score Fusion.}
Although the attribute and relation branches provide complementary evidence, anomalies in relational data are often localized to only a few semantic blocks or relation types. Directly averaging all residuals may dilute these local signals, while relying on a single branch may miss anomalies from other sources. To address this issue, \ourmethod adopts a two-level fusion of three local anomaly scores computed from normalized block- and relation-level errors:
\begin{equation}
\label{eq:final_score}
\begin{split}
s(u) = \;& \alpha\, s_{attr}(u) \\
& + (1{-}\alpha)\!\left[\beta\, s_{rel,self}(u) + (1{-}\beta)\, s_{rel,child}(u)\right],
\end{split}
\end{equation}
\noindent where $\alpha\in[0,1]$ controls the trade-off between attribute-local anomalies and relational anomalies, and $\beta\in[0,1]$ controls the trade-off between the self-profile and child-profile relational branches. This hierarchical parameterization reduces the fusion space while preserving interpretability: the attribute branch dominates when anomalies appear as local profile deviations, while the two relational branches provide complementary evidence when anomalies are reflected by inconsistent multi-relational behaviors. The algorithmic details and complexity analysis of \ourmethod are provided in Appendices~\ref{app:algo} and~\ref{app:complex}, respectively.

%% file: Tables/main_result.tex
\begin{table*}[t]
\centering
\resizebox{\textwidth}{!}{%
\renewcommand{\arraystretch}{1.15}
\begin{tabular}{l|cc|cc|cc|cc|cc|cc}
\toprule
\multirow{2}{*}{\textbf{Methods}}
& \multicolumn{2}{c|}{\textbf{Amazon}}
& \multicolumn{2}{c|}{\textbf{ArXiv}}
& \multicolumn{2}{c|}{\textbf{Avito}}
& \multicolumn{2}{c|}{\textbf{HM}}
& \multicolumn{2}{c|}{\textbf{SALT}}
& \multicolumn{2}{c}{\textbf{Stack}} \\
\cline{2-13}
& AUROC & AUPRC
& AUROC & AUPRC
& AUROC & AUPRC
& AUROC & AUPRC
& AUROC & AUPRC
& AUROC & AUPRC \\
\midrule
\rowcolor{gray!15}
\multicolumn{13}{c}{\textbf{GAD Methods}} \\
PREM
& 48.80 & 4.92
& 45.87 & 4.21
& 63.08 & 5.23
& 46.82 & 4.34
& 54.45 & 7.10
& 50.27 & 5.10 \\
FreeGAD
& 49.61 & 5.43
& 46.70 & 4.21
& 33.49 & 1.96
& 56.60 & 5.33
& 33.69 & 3.44
& 52.11 & 5.18 \\
DOMINANT
& OOM & OOM
& OOM & OOM
& OOM & OOM
& OOM & OOM
& 63.54 & 8.36
& OOM & OOM \\
\midrule
\rowcolor{gray!15}
\multicolumn{13}{c}{\textbf{TAD Methods}} \\
MCMTAD
& 47.21 & 4.70
& 51.11 & 5.13
& 67.64 & 4.37
& 80.27 & 16.37
& 72.75 & 8.94
& 61.10 & 6.83 \\
DRL
& 67.40 & 9.67
& 48.27 & 4.55
& 71.20 & 5.25
& 46.34 & 4.26
& 46.13 & 4.86
& 48.00 & 4.70 \\
KNN
& 54.40 & 5.79
& 46.49 & 4.46
& 63.19 & 3.52
& 69.06 & 8.82
& 67.14 & 7.10
& 50.28 & 4.99 \\
LOF
& 51.62 & 5.57
& 49.88 & 4.86
& 48.58 & 2.85
& 49.01 & 4.96
& 50.70 & 5.08
& 49.96 & 5.00 \\
IsoForest
& 54.27 & 7.12
& 46.52 & 4.29
& 64.66 & 4.06
& 45.04 & 4.14
& 72.58 & 16.71
& 55.96 & 5.78 \\
LUNAR
& 50.00 & 4.99
& 51.38 & 5.08
& 50.46 & 2.89
& 49.66 & 4.97
& 50.00 & 5.00
& 50.02 & 5.06 \\
\midrule
\rowcolor{gray!15}
\multicolumn{13}{c}{\textbf{Proposed RAD Method}} \\
\textbf{\ourmethod}
& \textbf{74.37} & \textbf{11.77}
& \textbf{56.80} & \textbf{13.89}
& \textbf{72.32} & \textbf{5.59}
& \textbf{85.39} & \textbf{29.56}
& \textbf{84.90} & \textbf{17.22}
& \textbf{67.98} & \textbf{7.66} \\
\bottomrule
\end{tabular}}
\caption{Anomaly detection performance in terms of AUROC and AUPRC. OOM denotes out-of-memory on a 24GB GPU. Best performance is highlighted in \textbf{bold}.}
\label{tab:main_results}
\end{table*}

%% file: Sections/5_exp.tex
\subsection{Experiment Setup}

\noindent\textbf{Datasets.}
We conduct experiments on six public relational databases from Relbench~\cite{robinson2024relbench} and Relbench v2~\cite{gu2026relbench} across multiple domains, including 3 e-commerce platforms (Amazon, HM, and Avito), an academic citation network (ArXiv), a Q\&A community (Stack), and an enterprise resource planning system (SALT~\cite{klein2024salt}). Detailed statistics are provided in Appendix~\ref{app:datasets}. As these benchmarks lack ground-truth anomaly labels, we design dataset-specific anomaly injection rules for evaluation. We simulate plausible fraud scenarios grounded in each dataset's relational structure. For example, in Amazon we replace a subset of a user's reviews with products from unrelated categories to simulate review manipulation, while in ArXiv we redirect citation edges to shared beacon papers to simulate citation cartels. The anomaly ratio is set to 5\% across all datasets. Injection procedures are provided in Appendix~\ref{app:anomalyInjection}.

\noindent\textbf{Baselines.} We compare \ourmethod with the representative GAD and TAD methods. \ding{182} GAD methods include PREM~\cite{pan2023prem}, FreeGAD~\cite{zhao2025freegad} and DOMINANT~\cite{ding2019deep}. \ding{183} TAD methods include MCMTAD~\cite{yin2024mcm}, DRL~\cite{ye2025drl}, KNN~\cite{ramaswamy2000efficient}, LOF~\cite{breunig2000lof}, IsolationForest~\cite{liu2019generative} and LUNAR~\cite{goodge2022lunar}.

\noindent\textbf{Evaluation and Implementation.} We report AUROC and AUPRC as the main metrics. For all methods, we report the mean over 5 random seeds. In our implementation, we optimize \ourmethod using Adam with a learning rate of $5 \times 10^{-3}$, and train the model for 100 epochs with a batch size of 8192. More implementation details are given in Appendix~\ref{app:imp}.

\subsection{Main Results}
Table~\ref{tab:main_results} reports the comparison results in terms of AUROC and AUPRC (more results are in Appendix~\ref{app:auc_ap_append}). We have the following observations. \ding{182}~\ourmethod achieves the best AUROC on all six datasets, demonstrating its consistent effectiveness for RAD across diverse relational schemas and anomaly sources. Compared with the strongest baseline on each dataset, \ourmethod shows more than 10\% relative performance gain on 4 datasets. 
\ding{183}~\ourmethod also obtains the best AUPRC on all six datasets, with particularly large gains on ArXiv, HM, and SALT. This indicates that \ourmethod not only ranks anomalies higher overall, but also better identifies rare anomalous instances under the highly imbalanced detection setting. \ding{184}~Existing TAD methods can perform competitively on some datasets after flattening relational information into tabular features, such as DRL on Amazon and Avito and MCMTAD on HM and SALT. However, their performance varies significantly across datasets, suggesting that flattened features cannot consistently preserve relation-specific abnormal patterns. \ding{185}~GAD methods generally underperform and DOMINANT suffers from out-of-memory issues on most datasets. This confirms that directly adapting homogeneous graph anomaly detection methods is insufficient for RAD, where multiple typed foreign-key relations and heterogeneous instance attributes need to be modeled explicitly. Overall, these results validate the advantage of jointly modeling local attribute reconstruction and dual-view multi-relational edge reconstruction in \ourmethod.

\input{Tables/ablation}
\subsection{Ablation Study}
To verify the effectiveness of each component, we construct six variants of \ourmethod: \ding{182}~\textbf{w/o Attr.}, 
removing attribute reconstruction; \ding{183}~\textbf{w/o Relation}, removing multi-relational edge reconstruction; \ding{184}~\textbf{w/o Gating}, removing conditional sparse gates; \ding{185}~\textbf{w/o Block Dec.}, using one shared attribute decoder; 
\ding{186}~\textbf{w/o Self View}, removing the self-profile branch; and \ding{187}~\textbf{w/o Child View}, removing the child-profile branch. Table~\ref{tab:ablation} shows that all components are beneficial.
Relational reconstruction is the most critical component, confirming the importance of modeling abnormal connection patterns. 
Removing the attribute branch also hurts performance, indicating that local attribute deviations provide complementary evidence. The performance drops caused by \textbf{w/o Gating} and \textbf{w/o Block Dec.} further show that conditional feature selection and block-specific reconstruction reduce redundancy and preserve localized anomaly signals.
For the dual-view design, \textbf{w/o Child View} generally suffers a larger drop than \textbf{w/o Self View}, suggesting that child-table behavioral context is especially informative. The full model consistently outperforms both single-view variants, confirming that all designs are valid and effective. 
More results are in the Appendix~\ref{app:ablation_app}.

\subsection{Hyperparameter Analysis}
Fig.~\ref{fig:hyperparams} shows the sensitivity of \ourmethod to $\alpha$ and $\beta$ on Amazon and SALT (more results are in Appendix~\ref{app:hyper_app}). The importance of the two weights differs notably across datasets. SALT requires a higher $\alpha$, indicating that attribute deviations play a major role in its anomaly pattern; in contrast, Amazon requires a smaller $\alpha$, relying more on multi-relational edge reconstruction. 
\input{Figures/HP_AlphaBeta/HP_alphabeta}
For $\beta$, most datasets prefer moderate values, confirming that both relational views provide complementary evidence. This disparity suggests that anomalies in different relational databases may manifest primarily in attributes or in relational connections, and \ourmethod can adapt the fusion weights to the underlying anomaly source.

\begin{figure}[t]
    \centering
    \includegraphics[width=0.48\textwidth]{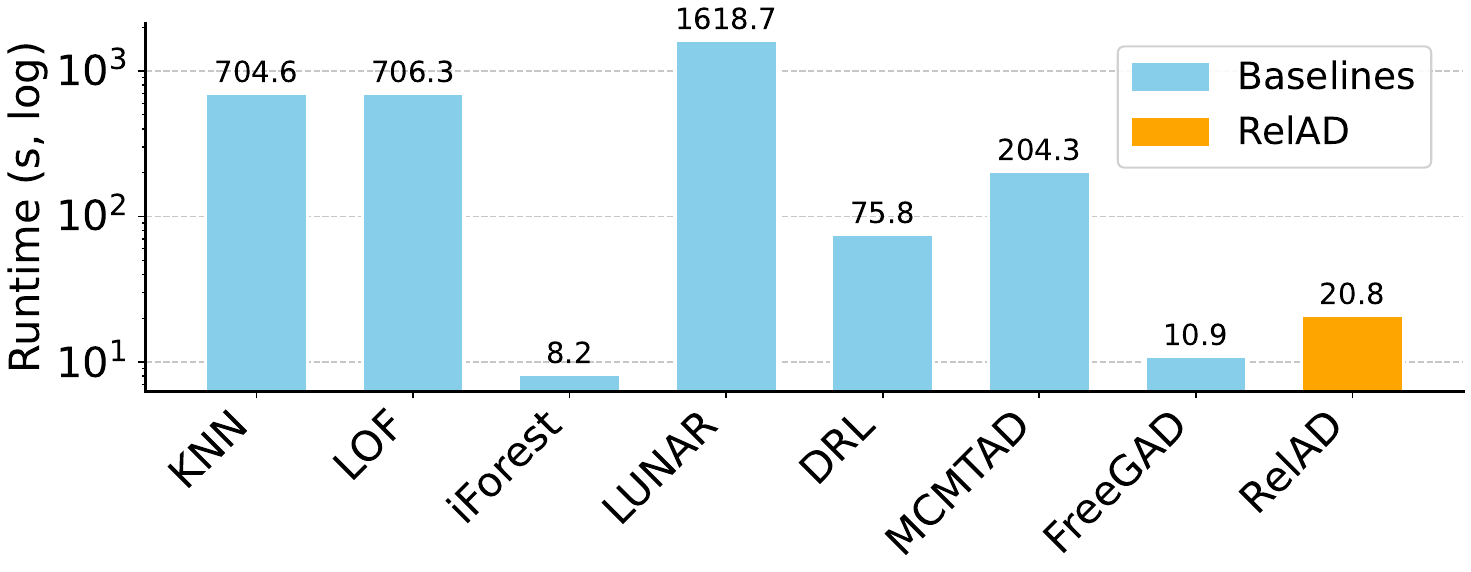}
    \caption{Runtime comparison on the HM dataset.}
    \label{fig:efficiency}
\end{figure}

\subsection{Efficiency Analysis}
To assess the runtime efficiency of \ourmethod, we compare its total running time with representative baselines on the HM dataset, excluding methods that run OOM. For a fair comparison, all trainable methods are evaluated under the same 5-epoch setting. As shown in Fig.~\ref{fig:efficiency}, \ourmethod achieves competitive efficiency while maintaining strong detection performance. In particular, \ourmethod is substantially faster than several costly baselines such as KNN, LOF, LUNAR, DRL, and MCMTAD, and its runtime remains close to the fastest methods. These results indicate that \ourmethod introduces limited computational overhead and remains practical for relational anomaly detection on large-scale datasets.

%% file: Tables/ablation.tex
\begin{table}[t]
\centering
\resizebox{\columnwidth}{!}{
\begin{tabular}{l|cccccc}
\toprule
Variant
& \textbf{Amazon}
& \textbf{ArXiv}
& \textbf{Avito}
& \textbf{HM}
& \textbf{SALT}
& \textbf{Stack} \\
\midrule
\ourmethod
& \textbf{74.37}
& \textbf{56.80}
& \textbf{72.32}
& \textbf{85.39}
& \textbf{84.90}
& \textbf{67.98} \\
\midrule
w/o Attr.
& 72.71
& 56.12
& 66.90
& 85.01
& 65.74
& 54.24 \\
w/o Relation
& 54.61
& 46.73
& 71.85
& 66.84
& 80.80
& 67.52 \\
w/o Gating
& 72.59
& 56.60
& 70.58
& 80.53
& 81.48
& 66.66 \\
w/o Block Dec.
& 71.82
& 55.53
& 67.38
& 84.42
& 81.80
& 63.06 \\
\midrule
w/o Self View
& 70.83
& 50.25
& 68.64
& 82.58
& 83.37
& 67.40 \\
w/o Child View
& 67.29
& 54.97
& 71.79
& 68.11
& 83.11
& 67.47 \\
\bottomrule
\end{tabular}}
\caption{AUROC of \ourmethod and its variants across six datasets.}
\label{tab:ablation}
\end{table}

%% file: Figures/HP_AlphaBeta/HP_alphabeta.tex
\begin{figure}[t]
    \centering
    \vspace{-4mm} 
    \subfigure[Amazon\label{subfig:hp_amazon}]{
        \includegraphics[width=0.46\columnwidth]{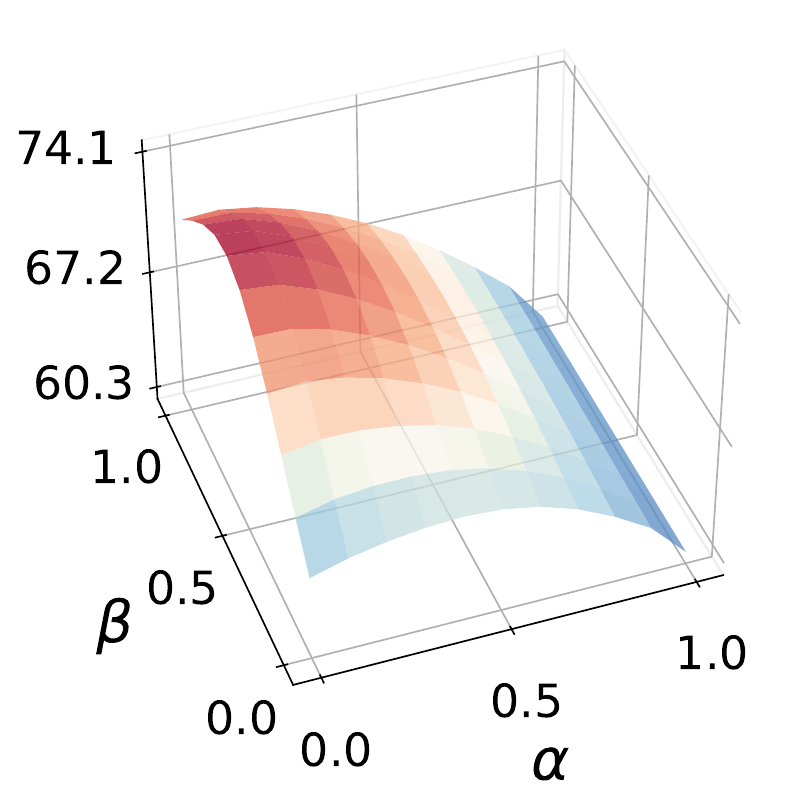}
    }\hfill
    \subfigure[SALT\label{subfig:hp_salt}]{
        \includegraphics[width=0.46\columnwidth]{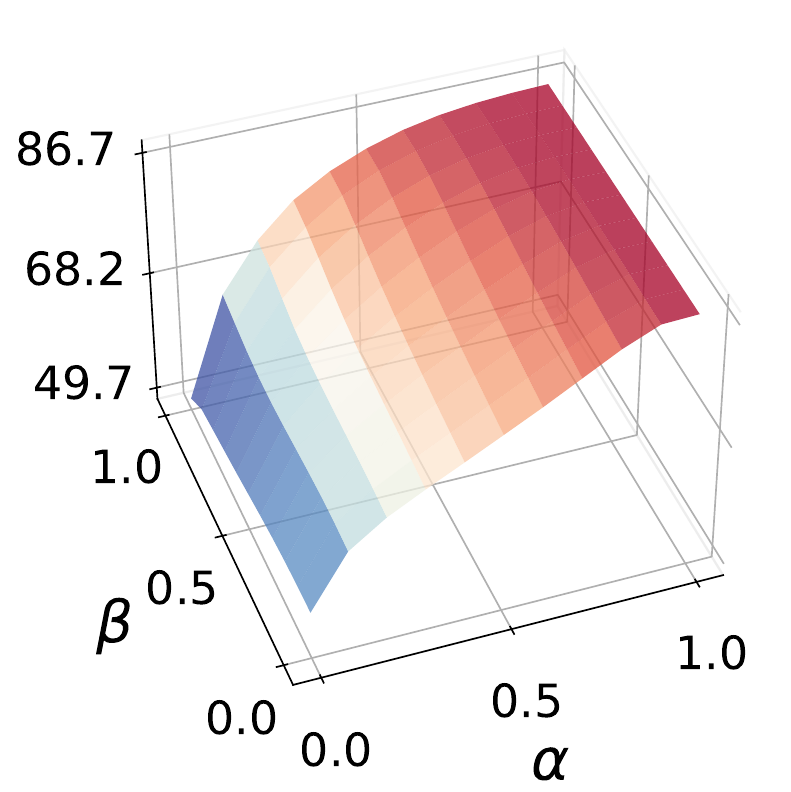}
    }
    \caption{Sensitivity w.r.t. $\alpha$ and $\beta$.}
    \label{fig:hyperparams}
\end{figure}

%% file: Sections/6_con.tex
In this paper, we take the first step towards addressing the relational anomaly detection (RAD) problem, aiming to identify abnormal instances from heterogeneous multi-table attributes and typed foreign-key relations. We introduce \ourmethod, a reconstruction-based RAD framework that jointly captures local attribute deviations and relation-specific abnormal connection patterns via attribute and edge reconstruction. Extensive experiments on six relational benchmark datasets demonstrate the detection prowess, robustness, and efficiency of \ourmethod compared to representative tabular and graph anomaly detection approaches. One \textbf{limitation} is that the current setting does not consider the temporal dimension and its influence on anomaly formation. A potential \textbf{future direction} is to extend \ourmethod to dynamic relational databases where instances, attributes, and foreign-key relations evolve over time.

%% file: Sections/7_appendix.tex
\section{Detailed Related Work} \label{app:rw}
\subsection{Relational Deep Learning}
Relational deep learning (RDL) aims to learn from multi-table relational databases by modeling instances and their primary--foreign key relationships~\cite{robinson2024relational,dwivedi2025relationalGT,chen2025relgnn,schlichtkrull2018modeling,miao2026blindguard,qian2026dynhd}. To advance this field, prior works~\cite{robinson2024relbench,gu2026relbench,wu2025large} have established large-scale relational benchmarks with diverse real-world datasets and tasks, thereby promoting the development of relational foundation models and representation learning methods. Among them, Griffin~\cite{wang2025griffin} transforms relational databases into heterogeneous graphs for graph-centric relational representation learning. RT~\cite{ranjan2025relational} introduces a Transformer-based framework to model rows, columns, and inter-table relations through relational attention. PLUREL~\cite{kothapalli2026plurel} further studies scaling laws for relational foundation models and shows that large-scale synthetic relational data can significantly improve relational pretraining and generalization.

Despite these advances, existing RDL studies mainly focus on predictive tasks such as classification, recommendation, and representation learning, while anomaly detection on relational data remains unexplored. In this work, we formally define the problem of relational anomaly detection, aiming to identify abnormal central instances from relational databases with multiple related tables and foreign-key dependencies, and construct corresponding benchmark datasets and evaluation protocols.

\subsection{Tabular Anomaly Detection}
Tabular anomaly detection (TAD) aims to detect anomalous instances by measuring deviations from normal patterns in structured tabular data~\cite{zuo2025rethinking,pang2021deep,borisov2022deep,shenkar2022anomaly,zuo2023sucola}. Classical methods rely on fixed inductive biases in the original feature space, including isolation-based approaches such as Isolation Forest~\cite{liu2019generative}, density-based methods such as LOF~\cite{breunig2000lof}, and distance-based methods such as kNN scoring~\cite{ramaswamy2000efficient}. These methods are computationally efficient but rely heavily on handcrafted assumptions about data geometry, which limits their robustness in complex data distributions. Recent advances in deep learning-based TAD have achieved strong performance improvements~\cite{thimonier2024retrieval,tsai2025anollm,ye2025disentangling,chang2023data}. Among them, representation learning-based methods form one of the most effective directions. For instance, MCM~\cite{yin2024mcm} introduces masked cell modeling to reconstruct randomly masked feature values, enabling the model to learn feature interactions in a self-supervised manner. DRL~\cite{ye2025drl} further proposes decomposed representation learning to disentangle latent factors and enhance anomaly separability. Despite their effectiveness, existing TAD methods are fundamentally designed for single-table data and rely on learning from flattened feature representations. When applied to relational data, they require aggregating multiple related tables into a single feature vector, which inevitably destroys interaction semantics encoded by foreign-key relations and mixes heterogeneous features in a unified space.

\subsection{Graph Anomaly Detection}
Graph anomaly detection (GAD) aims to identify anomalous data that are different from the majority~\cite{ma2021comprehensive,qiao2025deep,ho2025graph,yan2025address}, involving node, edge, and graph levels. In this paper, we focus on node-level anomaly detection on graphs~\cite{peng2018anomalous,li2017radar,dong2025smoothgnn}. Recent advances utilize graph neural networks (GNNs) to build powerful GAD models~\cite{qiao2025deep,pan2023prem,qiao2023truncated,chen2025uncertainty,tan2024influence}. For instance, DOMINANT~\cite{ding2019deep} introduces GCN-based graph autoencoder to reconstruct both structure and node attribute information. CoLA~\cite{liu2021anomaly} utilizes a contrastive GNN framework, combining contrastive learning and subgraph extraction to perform self-supervised GAD. Although these GAD methods have shown efficacy on homogeneous graphs, real-world relational data typically involves heterogeneous instance associations~\cite{bing2023heterogeneous}, e.g., user–product interactions in risk-control systems. While collapsing heterogeneous relations into a homogeneous graph can address this issue, it may discard relation-specific semantics. Therefore, exploiting heterogeneous instance relationships for anomaly detection remains an open challenge.

\section{Algorithm Description} \label{app:algo}
The training and testing algorithms are given in Algorithm~\ref{algo:train} and Algorithm~\ref{algo:test}, respectively.
\input{Tables/algorithm}

\section{Complexity Analysis} \label{app:complex}
In this subsection, we discuss the time complexity of \ourmethod in the testing phase. The overall cost mainly consists of four stages: conditional sparse-gated attribute reconstruction, dual-view instance encoding, dual-view multi-relational scoring, and hierarchical anomaly score fusion. The complexity of conditional sparse-gated attribute reconstruction is $\mathcal{O}\big(N(d_0h+\sum_{t=1}^{B}(d_0+d_t)h+dh+hd_z+\sum_{t=0}^{B}(d_zh+hd_t)+(B+1)\log K)\big)$, where $N$ is the number of target instances, $B$ is the number of child-table aggregated blocks in Eq.~\eqref{eq:gated_input}, $d_t$ is the dimension of the $t$-th block, and $d=\sum_{t=0}^{B}d_t$. This term covers block-wise gate generation, attribute encoding, block-specific decoding, and top-$K$ block selection in Eq.~\eqref{eq:attr_topk}. The complexity of dual-view instance encoding is $\mathcal{O}\big(N(d_0h+hd_{rel}+d_{child}h+hd_{rel})\big)$, where $d_{child}$ is the dimension of $\mathbf{x}_{u}^{child}$. The complexity of dual-view multi-relational scoring is $\mathcal{O}\big(\sum_{r\in\mathcal{R}} |\mathcal{V}^r|(d_rh+hd_{rel}) + d_{rel}\sum_{r\in\mathcal{R}}|\mathcal{E}_r| + N|\mathcal{R}|\log K\big)$, where $|\mathcal{V}^r|$ and $|\mathcal{E}_r|$ denote the number of neighbor instances and positive edges under relation $r$, respectively, and $d_r$ is the feature dimension of neighbor instances under relation $r$. The first term encodes relation-specific neighbor instances in Eq.~\eqref{eq:neighbor_entity_encoders}, the second term computes positive-edge likelihoods for the two branches up to a constant factor, and the last term selects the top-$K$ relation-wise scores in Eq.~\eqref{eq:rel_topk}. The complexity of hierarchical anomaly score fusion in Eq.~\eqref{eq:final_score} is $\mathcal{O}(N)$.

In summary, by treating the feature dimensions, hidden dimensions, the number of child-table aggregated blocks $B$, the number of relation types $|\mathcal{R}|$, and the latent dimension $d_{rel}$ as constants, the overall testing complexity simplifies to $\mathcal{O}(N + \sum_{r\in\mathcal{R}} |\mathcal{V}^r| + \sum_{r\in\mathcal{R}} |\mathcal{E}_r|)$. This demonstrates that the inference time of \ourmethod scales linearly with the total number of target instances, neighbor instances, and relation-specific edges, ensuring its practical efficiency and scalability for large-scale relational databases.

\section{Datasets} \label{app:datasets}
\input{Tables/app_dataset}

To comprehensively evaluate the generalization capability and practical effectiveness of our proposed model, we introduce six large-scale, real-world relational datasets into our experiments. These datasets span multiple core domains, including e-commerce, question-and-answer (Q\&A) communities, fashion retail, online advertising, enterprise resource planning (ERP) systems, and academic citation networks. They encompass massive heterogeneous instances and record rich multimodal attributes along with complex, dynamic interaction relations, providing a challenging and realistic evaluation environment to examine the model's performance in complex network topologies and real-world relational reasoning tasks. Dataset statistics are summarized in Table~\ref{tab:app_dataset}, with detailed descriptions provided below:

\begin{itemize}
    \item \textbf{Amazon:} This dataset focuses on the e-commerce interactions within the books category of the Amazon platform. It comprehensively records product instances (e.g., price, category), user profiles, and massive review interactions containing rich unstructured text, numerical ratings, and ``verified purchase'' labels. It provides multi-dimensional feature support for modeling the deep associations among user preferences, implicit intents, and product characteristics.
    
    \item \textbf{Stack:} This dataset originates from the Stats Stack Exchange site within the Stack Exchange Q\&A network, focusing on the domains of statistics and machine learning. It covers multi-dimensional user activity trajectories based on a community reputation system, including user biographies, Q\&A texts, edit histories, and voting behaviors. It captures the temporal evolution of information flow and self-moderating mechanisms within a knowledge community, providing abundant temporal network features driven by collective behavior.
    
    \item \textbf{HM:} This dataset integrates massive online transaction records and rich instance metadata from the globally renowned fast-fashion brand H\&M. It encompasses detailed customer demographic features and specific parameters for various product categories, exhibiting typical characteristics of long-term, large-scale consumer behavior. It serves as a high-quality benchmark for studying consumer purchasing patterns and product lifecycle predictions by delineating historical transaction relations within complex transaction graphs.
    
    \item \textbf{Avito:} This dataset originates from a contextual click-through rate (CTR) prediction competition hosted by Avito, a leading Russian classified advertisements platform. It encompasses large-scale user search query logs, detailed descriptions of ad attributes, and related exposure context features. It focuses on mining the precise matching relationships between search intents and ad contents, making it a representative scenario for evaluating a model's ability to process high-dimensional sparse features and complex contextual interactions.
    
    \item \textbf{SALT:} This dataset utilizes end-to-end real-world enterprise business transaction data released by SAP AI Research, collected directly from authentic Enterprise Resource Planning (ERP) systems. It covers sales document headers, line items, linked customer master data, and business workflow instances (e.g., sales offices, shipping points) with precise timestamps. It preserves an extremely high degree of industrial authenticity, focusing on the prediction of operational variables in practical supply chain and order fulfillment scenarios.
    
    \item \textbf{ArXiv:} This dataset constructs a large-scale academic literature network based on the arXiv physics domain between 2018 and 2023. It encompasses over 222,000 papers, 143,000 uniquely mapped authors via ORCID, 1.5 million directed citation links, and a hierarchical taxonomy of 53 physics subcategories. It provides a highly dense and complex many-to-many scientific knowledge graph, serving as a high-fidelity validation space for evaluating the model's ability to track scientific research evolution and analyze academic network topologies.
\end{itemize}

\section{Anomaly Injection Procedure} \label{app:anomalyInjection}
This section details the anomaly injection methods and implementation specifics applied across six relational datasets. To accurately evaluate the performance of anomaly detection algorithms in real-world business scenarios and effectively prevent models from relying on simple statistical features for ``shortcut learning,'' our anomaly injection process strictly adheres to three core design principles:

\begin{enumerate}
    \item \textbf{Stratified Injection Rate:} We initially sample 5\% of the total population across all datasets using a stratified sampling strategy (e.g., binning by node activity or degree) to eliminate selection bias, ensuring that the statistical distribution of the anomalous group rigorously aligns with that of the normal group. Importantly, anomaly labels are assigned only after verifying that the sampled instances satisfy the dataset-specific injection constraints. Consequently, while our target injection ratio is 5\%, the final proportion of labeled anomalies closely approximates this target but may be slightly lower in certain datasets due to these validity constraints.
    
    \item \textbf{Strict ``Replace-Only'' Strategy:} We strictly enforce a ``replace-only'' strategy during the injection phase, firmly prohibiting the addition or deletion of any data rows. This guarantees that the foundational statistical features of the nodes (e.g., interaction frequency, total degree) remain absolutely conserved before and after injection.
    
    \item \textbf{Real-World Scenario Reconstruction:} Each injection strategy is explicitly designed to reconstruct real-world business fraud scenarios, discarding mere random noise.
\end{enumerate}

Guided by these principles, we design a separate injection rule for each dataset according to its specific schema and business context. The per-dataset details are as follows:

\begin{itemize}
    \item \textbf{Amazon.}
    In e-commerce platforms, review manipulation (i.e., a user suddenly posting reviews on products in categories unrelated to their purchase history) is a common fraud pattern.
    We treat such category-shifting users as anomalous.
    Specifically, we select 5\% of users via stratified sampling from those whose review count exceeds the 20th-percentile threshold and whose dominant category share is above 25\%.
    For each selected user, we choose a target category different from their dominant one and replace 5--10 of their reviews with products from that category, resampling the rating and verified status from the target category distribution.

    \item \textbf{ArXiv.}
    In academic citation networks, citation cartels (i.e., a group of authors collectively citing a shared set of papers to inflate visibility) represent a realistic form of misconduct.
    We treat such coordinated citing authors as anomalous.
    Specifically, we sample 5\% of authors with at least one outgoing citation via degree-weighted sampling and organize them into cartels of 6--14 members.
    For each cartel, 15 high-citation papers spanning multiple categories are selected as shared beacons, and 40\% of each member's outgoing citation edges are replaced with these beacon papers.
 
    \item \textbf{Avito.}
    Click farming (i.e., a user concentrating clicks on a small set of paid ads to generate fake traffic) is a prevalent fraud scenario in online advertising.
    We treat users exhibiting such concentrated click patterns as anomalous.
    Specifically, we uniformly sample 5\% of users and replace 80\%--100\% of their clicked rows in the \textit{SearchStream} table with 5 fixed high-priced context ads, setting the position to 1--2.
    The \textit{VisitStream} and \textit{PhoneRequestsStream} tables are similarly redirected to the same target ad pool.
 
    \item \textbf{HM.}
    Considering that scalper-like purchasing is a notable risk in fashion retail, we treat users who buy products far outside their dominant style preference as anomalous.
    Specifically, 5\% of active customers are selected via stratified sampling, and 18\%--35\% of their transactions are replaced with products from a different style category, with price resampled accordingly.
    
    \item \textbf{SALT.} In enterprise order systems, anomalous transactions often exhibit rare attribute combinations that seldom co-occur under normal operations. 
    We treat customers whose orders contain such rare-combination patterns as anomalous. Specifically, we select 5\% of active customers via two-dimensional stratified sampling and replace 3\%--8\% of their order line attributes (product, category, plant, and shipping point) with rare values drawn from the bottom 20th percentile of the global distribution.
 
    \item \textbf{Stack.} In Q\&A platforms, cross-topic comment bots that post under unrelated topics are a representative form of spam behavior. We sample 5\% of users via stratified sampling and redirect approximately 60\% of their comments to posts under non-dominant topics, keeping the comment text unchanged.

\end{itemize}

\input{Tables/main_auroc}
\input{Tables/main_auprc}
\section{Implementation Details} \label{app:imp}
\noindent \textbf{Architecture, Training, and Inference Details.}
In our experiments, all encoders, including the attribute encoder $\operatorname{Enc}_{attr}$, the dual-view relational encoders ($\operatorname{Enc}_{self}$ and $\operatorname{Enc}_{child}$), and the neighbor-instance encoders $\operatorname{NodeEnc}^{r}$, along with the block-specific decoders $\operatorname{Dec}_t$, are implemented using two-layer Multilayer Perceptrons (MLPs) with ReLU activation. We use $h$ to denote the hidden dimension of the two-layer MLPs, and set $h=1024$ for the attribute encoder, relational encoders, gate generators, and block-specific decoders. The attribute latent dimension is denoted by $d_z$ and set to $64$, while the relational latent dimension is denoted by $d_{rel}$ and also set to $64$. For the neighbor-instance encoders $\operatorname{NodeEnc}^{r}$, the hidden dimension is set to $256$, with input dimensions dynamically adapted to the specific feature blocks of $\mathcal{V}^r$. To stabilize training, Layer Normalization is applied in the intermediate layers of $\operatorname{Enc}_{attr}$ and the relational encoders. For each dataset, we tune the key optimization and fusion hyperparameters by random search, including the learning rate, weight decay, batch size, training epochs, sparsity coefficient $\lambda_s$, and fusion coefficients $\alpha$ and $\beta$.

Regarding the objective function and inference, the model is optimized by the overall objective in Eq.~(\ref{eq:overall_loss}), where $\lambda_s$ controls the sparsity regularization for compact feature selection. To address GPU memory bottlenecks caused by large-scale relational neighbor sets, we restrict the maximum number of neighbor instances residing on the GPU to $500,000$, while implementing an edge sampling mechanism that limits samples to $200,000$ per relation type per epoch.

\input{Tables/ablation_append}
\noindent\textbf{Data Preprocessing.}
The data processing pipeline initiates by integrating anomaly labels $\mathbf{y}$ with the central table $\mathcal{T}^{0}$ to define the target instances $\mathcal{U}=\{u_1,\dots,u_N\}$. Global instance IDs are mapped to a contiguous local index space. Following the initial features definition, we use the RelBench-style preprocessing protocol to obtain row representations $\mathbf{h}_{v}^{k}$ for each table, with normalization applied to numerical attributes and unified encodings applied to heterogeneous raw columns.

To capture inter-table dependencies, we instantiate the heterogeneous graph $\mathcal{G}$ by resolving predefined primary--foreign key schemas. This process extracts clean, relation-specific edge lists $\mathcal{E}_{r}$ connecting central instances $u \in \mathcal{U}$ to their neighbors $i \in \mathcal{V}^{r}$. Following Eq.~\eqref{eq:x_attr}, we compute the aggregated feature blocks $\mathbf{x}_{u}^{agg,b}$ via two distinct modes: (1) direct aggregation of schema-connected child tables (calculating descriptive statistics such as mean, standard deviation, and counts); and (2) edge-mediated aggregation of neighbor feature matrices $\mathbf{H}^{r}$ along $\mathcal{E}_{r}$. 

Finally, for each target instance $u$, the self-feature $\mathbf{x}_{u}^{self}$ is concatenated with all aggregated blocks $\mathbf{x}_{u}^{agg,b}$ to form the unified attribute matrix $\mathbf{X} \in \mathbb{R}^{N \times d}$. The complete preprocessing module outputs $\mathbf{X}$, neighbor features $\mathbf{H}^{r}$, relation-specific edges $\mathcal{E}_{r}$, and the label vector $\mathbf{y} \in \{0,1\}^{N}$, providing a well-aligned and comprehensive input formulation for the downstream anomaly scoring function $f$.

\noindent\textbf{Computing infrastructures.} 
We implement the proposed method with Python 3.9 and PyTorch 2.8.0. The key dependencies include scikit-learn, Pandas, and Numpy. All experiments are conducted on a Ubuntu server with Intel Xeon Platinum 8352V CPU (16 vCPU) and NVIDIA RTX 4090 GPU (24GB) with CUDA 11.8.

\section{Supplemental Experiments} \label{app:supExp}

\subsection{Performance Comparison in Terms of AUROC and AUPRC}\label{app:auc_ap_append}
Tables~\ref{tab:auroc_results} and \ref{tab:auprc_results} further report the means and standard deviations over multiple runs. The relatively low variances demonstrate the robustness and stability of \ourmethod across different datasets. The results show that \ourmethod not only achieves consistently strong performance across all datasets, but also maintains relatively low variance, demonstrating its robustness and stability for relational anomaly detection.

\subsection{Detailed Results of Ablation Study}\label{app:ablation_app}
Table~\ref{tab:ablation_append_auprc} reports the full AUPRC ablation results across all six datasets, largely corroborating the conclusions drawn from the AUROC analysis. Overall, the complete \ourmethod framework achieves the best or highly competitive AUPRC scores across diverse scenarios. Specifically, relational modeling (\textbf{w/o Relation}) remains the most crucial component on Amazon, ArXiv, and HM, where its removal leads to severe AUPRC degradation (e.g., a massive drop from 29.56 to 7.65 on HM, and 13.89 to 4.33 on ArXiv). Conversely, mirroring the AUROC findings, the attribute reconstruction branch (\textbf{w/o Attr}) proves indispensable on SALT, Stack, and Avito, causing the sharpest AUPRC drops when removed (e.g., 17.22 to 10.32 on SALT, and 7.66 to 5.63 on Stack).  Furthermore, conditional gating and block-specific decoding (\textbf{w/o Gating} and \textbf{w/o Block Dec.}) consistently improve AUPRC, validating their capability to isolate fine-grained anomaly signals. Regarding the dual-view design, both views provide complementary benefits depending on the dataset's structural semantics: removing the child view causes a stark drop on HM (29.56 to 9.77), whereas removing the self view is more detrimental on ArXiv (13.89 to 5.11). Although partial variants occasionally exhibit marginal AUPRC fluctuations on specific datasets (e.g., \textbf{w/o Relation} on SALT or \textbf{w/o Self View} on Stack), these minor variances are typical in highly imbalanced datasets. These results further confirm that \ourmethod benefits from all proposed components across diverse relational databases, maintaining the most robust and comprehensive performance globally.

\input{Figures/HP_AlphaBeta/HP_alphabeta_append}
\subsection{Detailed Results of Hyperparameter Analysis}\label{app:hyper_app}
Fig.~\ref{fig:hyperparams_app} further presents the sensitivity analysis of \ourmethod to $\alpha$ and $\beta$ on four more representative datasets, including ArXiv, Avito, HM and Stack, to verify the generalization of our model's parameter adaptability. Consistent with the observations on Amazon and SALT, the optimal settings of the two weights show distinct differences across datasets, which are highly coupled with the core anomaly patterns of the data. Specifically, ArXiv and HM achieve the optimal detection performance when $\alpha$ is set to a small value near 0, indicating that anomalies in these two datasets are mainly embodied in abnormal multi-relational connection patterns, so the model relies more on the multi-relational edge reconstruction module to identify anomalies. On the contrary, Avito and Stack obtain the best performance with a relatively large $\alpha$, which demonstrates that attribute deviations are the dominant manifestation of anomalies in these datasets, and the attribute deviation modeling module plays a more critical role in anomaly scoring.

For the balance weight $\beta$, all four datasets reach the peak performance at moderate values, rather than extreme settings of 0 or 1. This result is consistent with our previous findings, which further confirms that the two relational views modeled in \ourmethod can provide complementary anomaly evidence, and the moderate fusion of the two views can fully mine the relational information in the data to boost detection performance. The extended experimental results on more datasets consistently validate that anomalies in different relational databases may originate from either attribute deviations or relational connection abnormalities. Benefiting from the flexible weight tuning mechanism, \ourmethod can adaptively adjust the fusion weights according to the underlying anomaly source of different data, thus maintaining stable and superior detection performance across diverse application scenarios.

%% file: Tables/algorithm.tex
\begin{algorithm*}[ht]
\caption{The Training Algorithm of \ourmethod}
\label{algo:train}
\LinesNumbered
\KwIn{Attribute matrix $\mathbf{X}\in\mathbb{R}^{N\times d}$ with attribute blocks $\{\mathbf{x}^{(t)}\}_{t=0}^{B}$;
relation set $\mathcal{R}$ with edge sets $\{\mathcal{E}_r\}$ and neighbor features $\{\mathbf{H}^{r}\}$;
training epochs $E$; batch size $b$; sampled edges per relation $M$;
learning rate $\eta$; sparsity weight $\lambda_s$.}
\KwOut{Trained model parameters $\Theta=\{\operatorname{MLP}_t,\operatorname{Enc}_{attr},\operatorname{Dec}_t,\operatorname{Enc}_{self},\operatorname{Enc}_{child},\operatorname{NodeEnc}_{r}\}_{t=0}^{B,r\in\mathcal{R}}$.}

Apply feature scaling on $\mathbf{X}$ and on each $\mathbf{H}^{r}$\;
Initialize all model parameters $\Theta$\;
\For{$\mathit{epoch}=1$ \KwTo $E$}{
    Generate target-instance permutation $\pi$ over $\{1,\dots,N\}$\;
    \tcp{Pre-sample at most $M$ edges for each relation}
    \For{$r\in\mathcal{R}$}{
        $\tilde{\mathcal{E}}_r \leftarrow$ sample $\min(M,|\mathcal{E}_r|)$ edges from $\mathcal{E}_r$ uniformly\;
    }
    \For{each mini-batch $\mathcal{B}\subset\pi$ of size $b$}{
        \tcp{Step 1: Conditional Sparse-Gated Attribute Reconstruction}
        Compute block-wise masks $\{\mathbf{m}_t\}$ via Eq.~\eqref{eq:mask_child} and the gated input $\tilde{\mathbf{x}}$ via Eq.~\eqref{eq:gated_input}\;
        Encode $\mathbf{z}_{attr}=\operatorname{Enc}_{attr}(\tilde{\mathbf{x}})$ and reconstruct $\{\hat{\mathbf{x}}^{(t)}\}_{t=0}^{B}$ via Eq.~\eqref{eq:block_decoder}\;
        Compute $\mathcal{L}_{rec}$ via Eq.~\eqref{eq:attr_rec_loss} and $\mathcal{L}_{sparse}$ via Eq.~\eqref{eq:sparse_loss}\;

        \tcp{Step 2: Dual-View Multi-Relational Edge Reconstruction}
        Compute $\mathbf{z}_{u,self}^{rel}$ via Eq.~\eqref{eq:rel_self_encoder} and $\mathbf{z}_{u,child}^{rel}$ via Eq.~\eqref{eq:rel_child_encoder}\;
        Identify active relations $\mathcal{R}_{act}$ with valid sampled edges in $\{\tilde{\mathcal{E}}_r\}_{r\in\mathcal{R}}$ connected to instances in batch $\mathcal{B}$\;
        \For{$r\in\mathcal{R}_{act}$}{
            Sample a negative neighbor $j\sim\operatorname{Uniform}(\mathcal{V}^{r})$ for each in-batch positive edge $(u,i)\in\tilde{\mathcal{E}}_r$\;
            Encode shared neighbor embeddings $\mathbf{e}_{i}^{r},\mathbf{e}_{j}^{r}$ via Eq.~\eqref{eq:neighbor_entity_encoders}\;
            Compute $\mathcal{L}_{q}^{r}$ for $q\in\{self,child\}$ via Eq.~\eqref{eq:rel_branch_loss}\;
        }
        
        \tcp{Step 3: Objective Computation and Parameter Update}
        Compute the total objective $\mathcal{L}$ via Eq.~\eqref{eq:overall_loss}\;
        Update $\Theta$ using Adam: $\Theta\leftarrow \Theta-\eta\nabla_{\Theta}\mathcal{L}$\;
    }
}
\Return $\Theta$\;
\end{algorithm*}

\begin{algorithm*}[ht]
\caption{The Inference Algorithm of \ourmethod}
\label{algo:test}
\LinesNumbered
\KwIn{Attribute matrix $\mathbf{X}\in\mathbb{R}^{N\times d}$ with attribute blocks $\{\mathbf{x}^{(t)}\}_{t=0}^{B}$;
relation set $\mathcal{R}$ with edge sets $\{\mathcal{E}_r\}$ and neighbor features $\{\mathbf{H}^{r}\}$;
trained model parameters $\Theta$; batch size $b$; top-$K$ for local fusion; weights $\alpha,\beta\in[0,1]$.}
\KwOut{Final anomaly scores $\{s(u)\}_{u=1}^{N}$.}

\tcp{Stage 1: Forward Pass for Instance Representations and Attribute Reconstruction}
\For{each mini-batch $\mathcal{B}\subset\{1,\dots,N\}$ of size $b$}{
    Obtain the gated input $\tilde{\mathbf{x}}$, attribute latent $\mathbf{z}_{attr}$, and block reconstructions $\{\hat{\mathbf{x}}^{(t)}\}_{t=0}^{B}$ via Eq.~\eqref{eq:mask_child}--\eqref{eq:block_decoder}\;
    Compute and cache the block-level reconstruction errors $\mathcal{L}_t(u)$ for $u\in\mathcal{B}$ via Eq.~\eqref{eq:block_error}\;
    Encode and cache the relational representations $\mathbf{z}_{u,self}^{rel}$ and $\mathbf{z}_{u,child}^{rel}$ via Eq.~\eqref{eq:rel_self_encoder}--\eqref{eq:rel_child_encoder}\;
}

\tcp{Stage 2: Attribute-Level Local Anomaly Scoring}
Compute the local attribute anomaly score $s_{attr}(u)$ for all instances using the normalized top-$K$ block errors via Eq.~\eqref{eq:attr_topk}\;

\tcp{Stage 3: Dual-View Multi-Relational Anomaly Scoring}
\For{$r\in\mathcal{R}$}{
    Encode all neighbor instances $\{\mathbf{e}_{i}^{r}\}_{i\in\mathcal{V}^{r}}$ using the shared neighbor encoder via Eq.~\eqref{eq:neighbor_entity_encoders}\;
    Compute the average positive-edge negative log-likelihoods $\bar{\ell}_{self}^{r}(u)$ and $\bar{\ell}_{child}^{r}(u)$ for instances connected under relation $r$\;
}
Aggregate the dual-view relational anomaly scores $s_{rel,self}(u)$ and $s_{rel,child}(u)$ for all instances via Eq.~\eqref{eq:rel_topk}\;

\tcp{Stage 4: Hierarchical Anomaly Score Fusion}
Compute the final anomaly score $s(u)$ via Eq.~\eqref{eq:final_score}\;
\Return $\{s(u)\}_{u=1}^{N}$\;
\end{algorithm*}

%% file: Tables/app_dataset.tex
\begin{table*}[t]
\centering
\begin{tabular}{l|c|ccc|cccc}
\toprule

\multirow{2}{*}{Dataset}
& \multirow{2}{*}{Domain}
& \multicolumn{3}{c|}{Relation Database}
& \multicolumn{4}{c}{Anomaly Injection} \\

\cmidrule(lr){3-5}
\cmidrule(lr){6-9}

&
&
\#Tables
& \#Rows
& \#Cols
& \#Samples
& \#Normal
& \#Anomaly
& \%Anomaly \\

\midrule

Amazon
& E-commerce
& 3
& 15,000,713
& 15
& 1,850,193
& 1,759,500
& 90,693
& 4.9018\% \\

Stack
& Social
& 7
& 4,247,264
& 52
& 53,225
& 50,563
& 2,662
& 5.0014\% \\

HM
& E-commerce
& 3
& 16,664,809
& 37
& 999,345
& 949,377
& 49,968
& 5.0001\% \\

Avito
& E-commerce
& 8
& 20,679,117
& 42
& 98,250
& 95,526
& 2,724
& 2.7725\% \\

SALT
& Enterprise
& 4
& 4,257,145
& 31
& 14,710
& 13,974
& 736
& 5.0034\% \\

Arxiv
& Academic
& 6
& 2,146,112
& 21
& 123,967
& 117,929
& 6,038
& 4.8707\% \\

\bottomrule
\end{tabular}
\caption{Statistics of datasets.}\label{tab:app_dataset}
\end{table*}

%% file: Tables/main_auroc.tex
\begin{table*}[t]
\centering
\small
\renewcommand{\arraystretch}{1.15}
\begin{tabular}{l|cccccc}
\toprule
\textbf{Methods}
& \textbf{Amazon}
& \textbf{ArXiv}
& \textbf{Avito}
& \textbf{HM}
& \textbf{SALT}
& \textbf{Stack} \\
\midrule

\rowcolor{gray!15}
\multicolumn{7}{c}{\textbf{GAD Methods}} \\

PREM
& 48.80$_{\pm0.49}$
& 45.87$_{\pm0.19}$
& 63.08$_{\pm2.11}$
& 46.82$_{\pm0.67}$
& 54.45$_{\pm2.46}$
& 50.27$_{\pm0.67}$ \\

FreeGAD
& 49.61$_{\pm0.00}$
& 46.70$_{\pm0.00}$
& 33.49$_{\pm0.00}$
& 56.60$_{\pm0.00}$
& 33.69$_{\pm0.00}$
& 52.11$_{\pm0.00}$ \\

DOMINANT
& OOM
& OOM
& OOM
& OOM
& 63.54$_{\pm2.03}$
& OOM \\

\midrule

\rowcolor{gray!15}
\multicolumn{7}{c}{\textbf{TAD Methods}} \\

MCMTAD
& 47.21$_{\pm1.25}$
& 51.11$_{\pm0.14}$
& 67.64$_{\pm0.00}$
& 80.27$_{\pm0.12}$
& 72.75$_{\pm0.14}$
& 61.10$_{\pm0.11}$ \\

DRL
& 67.40$_{\pm2.98}$
& 48.27$_{\pm2.92}$
& 71.20$_{\pm2.37}$
& 46.34$_{\pm4.51}$
& 46.13$_{\pm6.17}$
& 48.00$_{\pm0.24}$ \\

KNN
& 54.40$_{\pm0.00}$
& 46.49$_{\pm0.00}$
& 63.19$_{\pm0.00}$
& 69.06$_{\pm0.00}$
& 67.14$_{\pm0.00}$
& 50.28$_{\pm0.00}$ \\

LOF
& 51.62$_{\pm0.00}$
& 49.88$_{\pm0.00}$
& 48.58$_{\pm0.00}$
& 49.01$_{\pm0.00}$
& 50.70$_{\pm0.00}$
& 49.96$_{\pm0.00}$ \\

IsoForest
& 54.27$_{\pm1.19}$
& 46.52$_{\pm0.13}$
& 64.66$_{\pm3.43}$
& 45.04$_{\pm0.08}$
& 72.58$_{\pm0.97}$
& 55.96$_{\pm2.06}$ \\

LUNAR
& 50.00$_{\pm0.00}$
& 51.38$_{\pm0.21}$
& 50.46$_{\pm0.45}$
& 49.66$_{\pm0.39}$
& 50.00$_{\pm0.00}$
& 50.02$_{\pm0.79}$ \\

\midrule

\rowcolor{gray!15}
\multicolumn{7}{c}{\textbf{Proposed RAD Method}} \\

\textbf{\ourmethod}
& \textbf{74.37$_{\pm0.36}$}
& \textbf{56.80$_{\pm0.19}$}
& \textbf{72.32$_{\pm0.77}$}
& \textbf{85.39$_{\pm1.03}$}
& \textbf{84.90$_{\pm0.31}$}
& \textbf{67.98$_{\pm0.79}$} \\

\bottomrule
\end{tabular}
\caption{Anomaly detection performance in terms of AUROC (in percent, mean±std). OOM denotes out-of-memory on a 24GB GPU. Best results are highlighted in \textbf{bold}.}\label{tab:auroc_results}
\end{table*}

%% file: Tables/main_auprc.tex
\begin{table*}[t]
\centering
\small
\renewcommand{\arraystretch}{1.15}
\begin{tabular}{l|cccccc}
\toprule
\textbf{Methods}
& \textbf{Amazon}
& \textbf{ArXiv}
& \textbf{Avito}
& \textbf{HM}
& \textbf{SALT}
& \textbf{Stack} \\
\midrule

\rowcolor{gray!15}
\multicolumn{7}{c}{\textbf{GAD Methods}} \\

PREM
& 4.92$_{\pm0.06}$
& 4.21$_{\pm0.01}$
& 5.23$_{\pm0.34}$
& 4.34$_{\pm0.06}$
& 7.10$_{\pm1.01}$
& 5.10$_{\pm0.05}$ \\

FreeGAD
& 5.43$_{\pm0.00}$
& 4.21$_{\pm0.00}$
& 1.96$_{\pm0.00}$
& 5.33$_{\pm0.00}$
& 3.44$_{\pm0.00}$
& 5.18$_{\pm0.00}$ \\

DOMINANT
& OOM
& OOM
& OOM
& OOM
& 8.36$_{\pm0.78}$
& OOM \\

\midrule

\rowcolor{gray!15}
\multicolumn{7}{c}{\textbf{TAD Methods}} \\

MCMTAD
& 4.70$_{\pm0.09}$
& 5.13$_{\pm0.00}$
& 4.37$_{\pm0.00}$
& 16.37$_{\pm0.14}$
& 8.94$_{\pm0.00}$
& 6.83$_{\pm0.00}$ \\

DRL
& 9.67$_{\pm0.21}$
& 4.55$_{\pm0.37}$
& 5.25$_{\pm0.45}$
& 4.26$_{\pm0.25}$
& 4.86$_{\pm0.18}$
& 4.70$_{\pm0.13}$ \\

KNN
& 5.79$_{\pm0.00}$
& 4.46$_{\pm0.00}$
& 3.52$_{\pm0.00}$
& 8.82$_{\pm0.00}$
& 7.10$_{\pm0.00}$
& 4.99$_{\pm0.00}$ \\

LOF
& 5.57$_{\pm0.00}$
& 4.86$_{\pm0.00}$
& 2.85$_{\pm0.00}$
& 4.96$_{\pm0.00}$
& 5.08$_{\pm0.00}$
& 5.00$_{\pm0.00}$ \\

IsoForest
& 7.12$_{\pm0.06}$
& 4.29$_{\pm0.01}$
& 4.06$_{\pm0.37}$
& 4.14$_{\pm0.04}$
& 16.71$_{\pm0.45}$
& 5.78$_{\pm0.32}$ \\

LUNAR
& 4.99$_{\pm0.00}$
& 5.08$_{\pm0.03}$
& 2.89$_{\pm0.04}$
& 4.97$_{\pm0.03}$
& 5.00$_{\pm0.00}$
& 5.06$_{\pm0.45}$ \\

\midrule

\rowcolor{gray!15}
\multicolumn{7}{c}{\textbf{Proposed RAD Method}} \\

\textbf{\ourmethod}
& \textbf{11.77$_{\pm0.25}$}
& \textbf{13.89$_{\pm0.23}$}
& \textbf{5.59$_{\pm0.20}$}
& \textbf{29.56$_{\pm2.04}$}
& \textbf{17.22$_{\pm0.55}$}
& \textbf{7.66$_{\pm0.15}$} \\

\bottomrule
\end{tabular}
\caption{Anomaly detection performance in terms of AUPRC (in percent, mean±std). OOM denotes out-of-memory on a 24GB GPU. Best results are highlighted in \textbf{bold}.}\label{tab:auprc_results}
\end{table*}

%% file: Tables/ablation_append.tex
\begin{table}[t]
\centering
\resizebox{\linewidth}{!}{%
\begin{tabular}{l|cccccc}
\toprule
Variant & Amazon & ArXiv & Avito & HM & SALT & Stack \\
\midrule
\ourmethod            & \textbf{11.77} & \textbf{13.89} & \textbf{5.59} & \textbf{29.56} & 17.22 & 7.66 \\
\midrule
w/o Attr         & 10.40 & 11.39 & 3.83 & 24.59 & 10.32 & 5.63 \\
w/o Relation     & 5.20 & 4.33 & 5.14 & 7.65 & \textbf{17.31} & 7.60 \\
w/o Gating       & 10.03 & 11.92 & 4.98 & 17.80 & 14.88 & 7.14 \\
w/o Block Dec    & 9.79 & 10.71 & 4.42 & 23.18 & 13.75 & 6.47 \\
\midrule
w/o Self View    & 9.71 & 5.11 & 4.65 & 22.75 & 16.00 & \textbf{7.72} \\
w/o Child View   & 9.04 & 9.88 & 5.31 & 9.77 & 16.59 & 7.54 \\
\bottomrule
\end{tabular}}
\caption{AUPRC of \ourmethod and its variants on all datasets.}\label{tab:ablation_append_auprc}
\end{table}

%% file: Figures/HP_AlphaBeta/HP_alphabeta_append.tex
\begin{figure}[!htbp]
    \centering
    \subfigure[ArXiv\label{subfig:hp_arxiv}]{
        \includegraphics[width=0.47\linewidth]{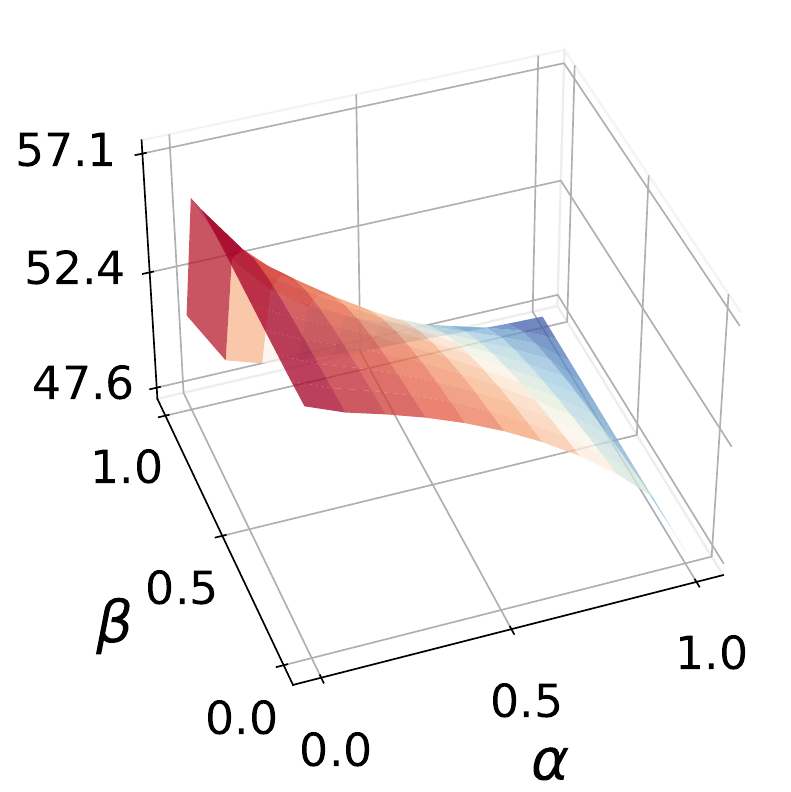}
    }
    \hfill
    \subfigure[Avito\label{subfig:hp_avito}]{
        \includegraphics[width=0.47\linewidth]{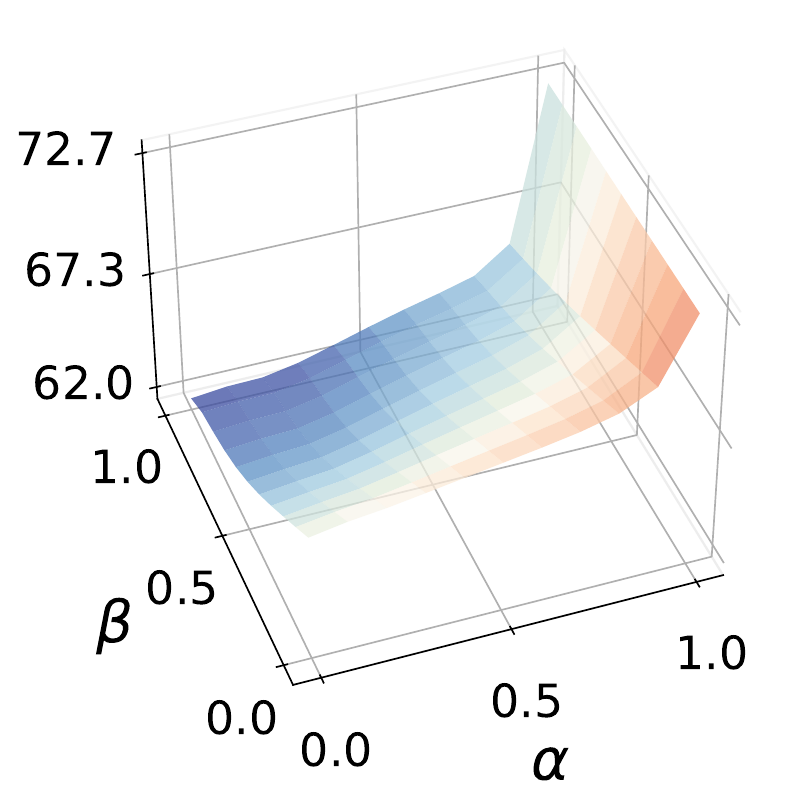}
    }
    \hfill
    \subfigure[HM\label{subfig:hp_hm}]{
        \includegraphics[width=0.47\linewidth]{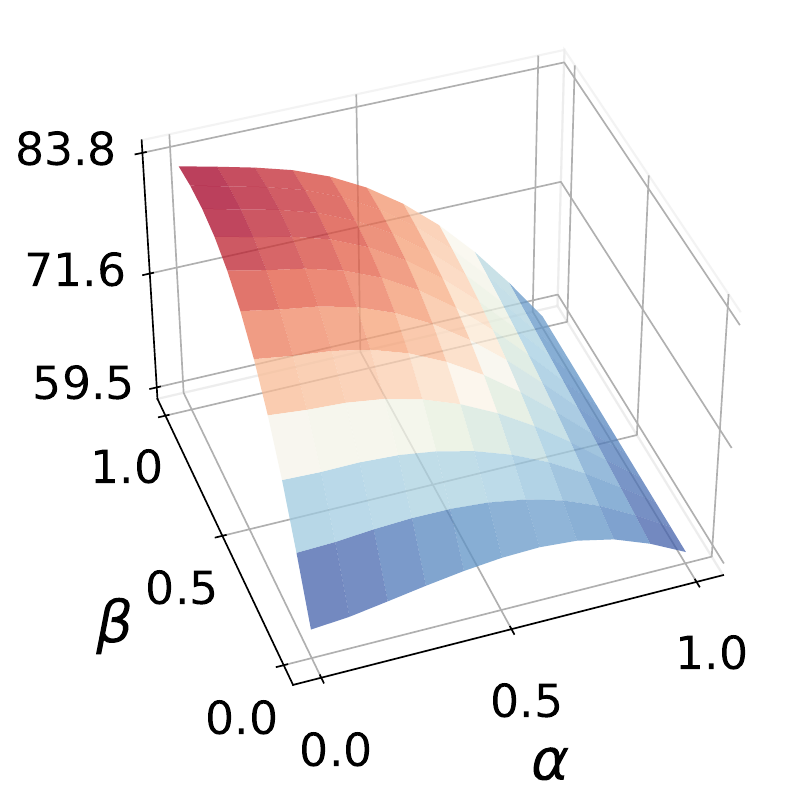}
    }
    \hfill
    \subfigure[Stack\label{subfig:hp_stack}]{
        \includegraphics[width=0.47\linewidth]{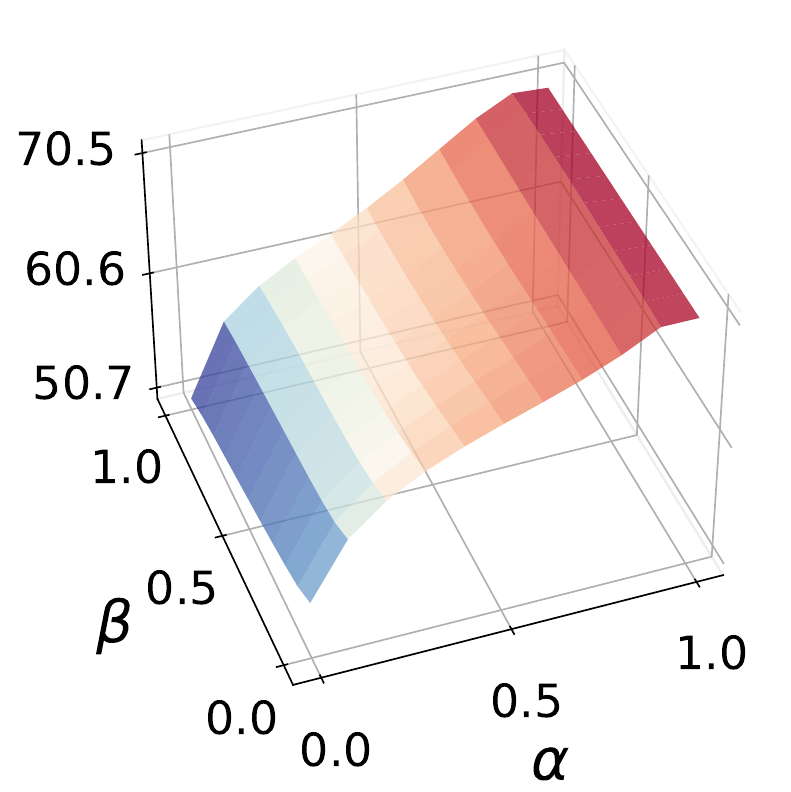}
    }

    \caption{Sensitivity analysis of \ourmethod with respect to $\alpha$ and $\beta$ on remaining datasets.}
    \label{fig:hyperparams_app}
\end{figure}

%% file: ref.bib
@article{robinson2024relbench,
  title={Relbench: A benchmark for deep learning on relational databases},
  author={Robinson, Joshua and Ranjan, Rishabh and Hu, Weihua and Huang, Kexin and Han, Jiaqi and Dobles, Alejandro and Fey, Matthias and Lenssen, Jan E and Yuan, Yiwen and Zhang, Zecheng and others},
  journal={Advances in Neural Information Processing Systems},
  volume={37},
  pages={21330--21341},
  year={2024}
}

@inproceedings{dwivedi2025relational,
  title={Relational deep learning: Challenges, foundations and next-generation architectures},
  author={Dwivedi, Vijay Prakash and Kanatsoulis, Charilaos and Huang, Shenyang and Leskovec, Jure},
  booktitle={Proceedings of the 31st ACM SIGKDD Conference on Knowledge Discovery and Data Mining V. 2},
  pages={5999--6009},
  year={2025}
}

@inproceedings{robinson2024relational,
  title={Relational deep learning: Graph representation learning on relational databases},
  author={Robinson, Joshua and Ranjan, Rishabh and Hu, Weihua and Huang, Kexin and Han, Jiaqi and Dobles, Alejandro and Fey, Matthias and Lenssen, Jan Eric and Yuan, Yiwen and Zhang, Zecheng and others},
  booktitle={NeurIPS 2024 third table representation learning workshop},
  year={2024}
}

@inproceedings{wu2025large,
  title={Large language models are good relational learners},
  author={Wu, Fang and Dwivedi, Vijay Prakash and Leskovec, Jure},
  booktitle={Proceedings of the 63rd Annual Meeting of the Association for Computational Linguistics (Volume 1: Long Papers)},
  pages={7835--7854},
  year={2025}
}

@article{pang2021deep,
  title={Deep learning for anomaly detection: A review},
  author={Pang, Guansong and Shen, Chunhua and Cao, Longbing and Hengel, Anton Van Den},
  journal={ACM computing surveys (CSUR)},
  volume={54},
  number={2},
  pages={1--38},
  year={2021},
  publisher={ACM New York, NY, USA}
}

@inproceedings{shenkar2022anomaly,
  title={Anomaly detection for tabular data with internal contrastive learning},
  author={Shenkar, Tom and Wolf, Lior},
  booktitle={International conference on learning representations},
  year={2022}
}

@article{ma2021comprehensive,
  title={A comprehensive survey on graph anomaly detection with deep learning},
  author={Ma, Xiaoxiao and Wu, Jia and Xue, Shan and Yang, Jian and Zhou, Chuan and Sheng, Quan Z and Xiong, Hui and Akoglu, Leman},
  journal={IEEE transactions on knowledge and data engineering},
  volume={35},
  number={12},
  pages={12012--12038},
  year={2021},
  publisher={IEEE}
}

@article{qiao2025deep,
  title={Deep graph anomaly detection: A survey and new perspectives},
  author={Qiao, Hezhe and Tong, Hanghang and An, Bo and King, Irwin and Aggarwal, Charu and Pang, Guansong},
  journal={IEEE Transactions on Knowledge and Data Engineering},
  year={2025},
  publisher={IEEE}
}

@inproceedings{li2017radar,
  title={Radar: residual analysis for anomaly detection in attributed networks},
  author={Li, Jundong and Dani, Harsh and Hu, Xia and Liu, Huan},
  booktitle={Proceedings of the 26th International Joint Conference on Artificial Intelligence},
  pages={2152--2158},
  year={2017}
}

@inproceedings{peng2018anomalous,
  title={ANOMALOUS: A joint modeling approach for anomaly detection on attributed networks.},
  author={Peng, Zhen and Luo, Minnan and Li, Jundong and Liu, Huan and Zheng, Qinghua and others},
  booktitle={Ijcai},
  volume={18},
  pages={3513--3519},
  year={2018}
}

@article{qiao2023truncated,
  title={Truncated affinity maximization: One-class homophily modeling for graph anomaly detection},
  author={Qiao, Hezhe and Pang, Guansong},
  journal={Advances in Neural Information Processing Systems},
  volume={36},
  pages={49490--49512},
  year={2023}
}

@inproceedings{dwivedi2025relationalGT,
  title = {Relational Graph Transformer},
  author = {Dwivedi, Vijay Prakash and Jaladi, Sri and Shen, Yangyi and Lopez, Federico and Kanatsoulis, Charilaos I. and Puri, Rishi and Fey, Matthias and Leskovec, Jure},
  booktitle = {Temporal Graph Learning Workshop @ KDD 2025},
  year      = {2025},
}

@inproceedings{chen2025relgnn,
  title={RelGNN: Composite Message Passing for Relational Deep Learning},
  author={Chen, Tianlang and Kanatsoulis, Charilaos and Leskovec, Jure},
  booktitle={International Conference on Machine Learning},
  pages={8296--8312},
  year={2025},
  organization={PMLR}
}

@article{gu2026relbench,
  title={RelBench v2: A Large-Scale Benchmark and Repository for Relational Data},
  author={Gu, Justin and Ranjan, Rishabh and Kanatsoulis, Charilaos and Tang, Haiming and Jurkovic, Martin and Hudovernik, Valter and Znidar, Mark and Chaturvedi, Pranshu and Shroff, Parth and Li, Fengyu and others},
  journal={arXiv preprint arXiv:2602.12606},
  year={2026}
}

@inproceedings{wang2025griffin,
  title={Griffin: Towards a Graph-Centric Relational Database Foundation Model},
  author={Wang, Yanbo and Wang, Xiyuan and Gan, Quan and Wang, Minjie and Yang, Qibin and Wipf, David and Zhang, Muhan},
  booktitle={International Conference on Machine Learning},
  pages={64604--64627},
  year={2025},
  organization={PMLR}
}

@inproceedings{ranjan2025relational,
  title={Relational Transformer: Toward Zero-Shot Foundation Models for Relational Data},
  author={Ranjan, Rishabh and Hudovernik, Valter and Znidar, Mark and Kanatsoulis, Charilaos I and Upendra, Roshan Reddy and Mohammadi, Mahmoud and Meyer, Joe and Palczewski, Tom and Guestrin, Carlos and Leskovec, Jure},
  booktitle={EurIPS 2025 Workshop: AI for Tabular Data},
  year = {2025}
}

@article{kothapalli2026plurel,
  title={PluRel: Synthetic Data unlocks Scaling Laws for Relational Foundation Models},
  author={Kothapalli, Vignesh and Ranjan, Rishabh and Hudovernik, Valter and Dwivedi, Vijay Prakash and Hoffart, Johannes and Guestrin, Carlos and Leskovec, Jure},
  journal={arXiv preprint arXiv:2602.04029},
  year={2026}
}

@article{liu2019generative,
  title={Generative adversarial active learning for unsupervised outlier detection},
  author={Liu, Yezheng and Li, Zhe and Zhou, Chong and Jiang, Yuanchun and Sun, Jianshan and Wang, Meng and He, Xiangnan},
  journal={IEEE Transactions on Knowledge and Data Engineering},
  volume={32},
  number={8},
  pages={1517--1528},
  year={2019},
  publisher={IEEE}
}

@inproceedings{breunig2000lof,
  title={LOF: identifying density-based local outliers},
  author={Breunig, Markus M and Kriegel, Hans-Peter and Ng, Raymond T and Sander, J{\"o}rg},
  booktitle={Proceedings of the 2000 ACM SIGMOD international conference on Management of data},
  pages={93--104},
  year={2000}
}

@inproceedings{ramaswamy2000efficient,
  title={Efficient algorithms for mining outliers from large data sets},
  author={Ramaswamy, Sridhar and Rastogi, Rajeev and Shim, Kyuseok},
  booktitle={Proceedings of the 2000 ACM SIGMOD international conference on Management of data},
  pages={427--438},
  year={2000}
}

@inproceedings{ding2019deep,
  title={Deep anomaly detection on attributed networks},
  author={Ding, Kaize and Li, Jundong and Bhanushali, Rohit and Liu, Huan},
  booktitle={Proceedings of the 2019 SIAM international conference on data mining},
  pages={594--602},
  year={2019},
  organization={SIAM}
}

@inproceedings{pan2023prem,
  title={PREM: A Simple Yet Effective Approach for Node-Level Graph Anomaly Detection},
  author={Pan, Junjun and Liu, Yixin and Zheng, Yizhen and Pan, Shirui},
  booktitle={2023 IEEE International Conference on Data Mining (ICDM)},
  pages={1253--1258},
  year={2023},
  organization={IEEE}
}

@article{liu2021anomaly,
  title={Anomaly detection on attributed networks via contrastive self-supervised learning},
  author={Liu, Yixin and Li, Zhao and Pan, Shirui and Gong, Chen and Zhou, Chuan and Karypis, George},
  journal={IEEE transactions on neural networks and learning systems},
  volume={33},
  number={6},
  pages={2378--2392},
  year={2021},
  publisher={IEEE}
}

@article{liu2024arc,
  title={Arc: A generalist graph anomaly detector with in-context learning},
  author={Liu, Yixin and Li, Shiyuan and Zheng, Yu and Chen, Qingfeng and Zhang, Chengqi and Pan, Shirui},
  journal={Advances in Neural Information Processing Systems},
  volume={37},
  pages={50772--50804},
  year={2024}
}

@inproceedings{zhao2025freegad,
  title={Freegad: A training-free yet effective approach for graph anomaly detection},
  author={Zhao, Yunfeng and Liu, Yixin and Li, Shiyuan and Chen, Qingfeng and Zheng, Yu and Pan, Shirui},
  booktitle={Proceedings of the 34th ACM International Conference on Information and Knowledge Management},
  pages={4379--4389},
  year={2025}
}

@inproceedings{goodge2022lunar,
  title={Lunar: Unifying local outlier detection methods via graph neural networks},
  author={Goodge, Adam and Hooi, Bryan and Ng, See-Kiong and Ng, Wee Siong},
  booktitle={Proceedings of the AAAI conference on artificial intelligence},
  volume={36},
  number={6},
  pages={6737--6745},
  year={2022}
}

@inproceedings{thimonier2024beyond,
  title={Beyond Individual Input for Deep Anomaly Detection on Tabular Data},
  author={Thimonier, Hugo and Popineau, Fabrice and Rimmel, Arpad and Doan, Bich-Li{\^e}n},
  booktitle={International Conference on Machine Learning},
  pages={48097--48123},
  year={2024},
  organization={PMLR}
}

@inproceedings{yin2024mcm,
  title={Mcm: Masked cell modeling for anomaly detection in tabular data},
  author={Yin, Jiaxin and Qiao, Yuanyuan and Zhou, Zitang and Wang, Xiangchao and Yang, Jie},
  booktitle={The Twelfth International Conference on Learning Representations},
  year={2024}
}

@inproceedings{ye2025drl,
  title={DRL: Decomposed representation learning for tabular anomaly detection},
  author={Ye, Hangting and Zhao, He and Fan, Wei and Zhou, Mingyuan and dan Guo, Dan and Chang, Yi},
  booktitle={The Thirteenth International Conference on Learning Representations},
  year={2025}
}

@inproceedings{chen2024boosting,
  title={Boosting graph anomaly detection with adaptive message passing},
  author={Chen, Jingyan and Zhu, Guanghui and Yuan, Chunfeng and Huang, Yihua},
  booktitle={The Twelfth International Conference on Learning Representations},
  year={2024}
}

@article{borisov2022deep,
  title={Deep neural networks and tabular data: A survey},
  author={Borisov, Vadim and Leemann, Tobias and Se{\ss}ler, Kathrin and Haug, Johannes and Pawelczyk, Martin and Kasneci, Gjergji},
  journal={IEEE transactions on neural networks and learning systems},
  volume={35},
  number={6},
  pages={7499--7519},
  year={2022},
  publisher={IEEE}
}

@inproceedings{klein2024salt,
  title={SALT: Sales Autocompletion Linked Business Tables Dataset},
  author={Klein, Tassilo and Biehl, Clemens and Costa, Margarida and Sres, Andre and Kolk, Jonas and Hoffart, Johannes},
  booktitle={NeurIPS 2024 Third Table Representation Learning Workshop},
  year={2024}
}

@inproceedings{schlichtkrull2018modeling,
  title={Modeling relational data with graph convolutional networks},
  author={Schlichtkrull, Michael and Kipf, Thomas N and Bloem, Peter and Van Den Berg, Rianne and Titov, Ivan and Welling, Max},
  booktitle={European semantic web conference},
  pages={593--607},
  year={2018},
  organization={Springer}
}

@inproceedings{thimonier2024retrieval,
  title={Retrieval augmented deep anomaly detection for tabular data},
  author={Thimonier, Hugo and Popineau, Fabrice and Rimmel, Arpad and Doan, Bich-Li{\^e}n},
  booktitle={Proceedings of the 33rd ACM international conference on information and knowledge management},
  pages={2250--2259},
  year={2024}
}

@inproceedings{tsai2025anollm,
  title={AnoLLM: Large language models for tabular anomaly detection},
  author={Tsai, Che-Ping and Teng, Ganyu and Wallis, Phillip and Ding, Wei},
  booktitle={The Thirteenth International Conference on Learning Representations},
  year={2025}
}

@inproceedings{ye2025disentangling,
  title={Disentangling tabular data towards better one-class anomaly detection},
  author={Ye, Jianan and Tan, Zhaorui and Hu, Yijie and Yang, Xi and Cheng, Guangliang and Huang, Kaizhu},
  booktitle={Proceedings of the AAAI Conference on Artificial Intelligence},
  volume={39},
  number={12},
  pages={13061--13068},
  year={2025}
}

@inproceedings{chang2023data,
  title={Data-efficient and interpretable tabular anomaly detection},
  author={Chang, Chun-Hao and Yoon, Jinsung and Arik, Sercan {\"O} and Udell, Madeleine and Pfister, Tomas},
  booktitle={Proceedings of the 29th ACM SIGKDD Conference on Knowledge Discovery and Data Mining},
  pages={190--201},
  year={2023}
}

@article{ho2025graph,
  title={Graph anomaly detection in time series: A survey},
  author={Ho, Thi Kieu Khanh and Karami, Ali and Armanfard, Narges},
  journal={IEEE Transactions on Pattern Analysis and Machine Intelligence},
  year={2025},
  publisher={IEEE}
}

@article{bing2023heterogeneous,
  title={Heterogeneous graph neural networks analysis: a survey of techniques, evaluations and applications.},
  author={Bing, Rui and Yuan, Guan and Zhu, Mu and Meng, Fanrong and Ma, Huifang and Qiao, Shaojie},
  journal={Artificial Intelligence Review},
  volume={56},
  number={8},
  year={2023}
}

@inproceedings{dong2025smoothgnn,
  title={Smoothgnn: Smoothing-aware gnn for unsupervised node anomaly detection},
  author={Dong, Xiangyu and Zhang, Xingyi and Sun, Yanni and Chen, Lei and Yuan, Mingxuan and Wang, Sibo},
  booktitle={Proceedings of the ACM on Web Conference 2025},
  pages={1225--1236},
  year={2025}
}

@inproceedings{li2026towards,
  title     = {Towards One-for-All Anomaly Detection for Tabular Data},
  author    = {Li, Shiyuan and Liu, Yixin and Zheng, Yu and Cao, Xiaofeng and Pan, Shirui and Shen, Heng Tao},
  booktitle = {Proceedings of the International Conference on Machine Learning},
  year      = {2026},
}

@inproceedings{pan2025label,
  title={A label-free heterophily-guided approach for unsupervised graph fraud detection},
  author={Pan, Junjun and Liu, Yixin and Zheng, Xin and Zheng, Yizhen and Liew, Alan Wee-Chung and Li, Fuyi and Pan, Shirui},
  booktitle={Proceedings of the AAAI Conference on Artificial Intelligence},
  volume={39},
  number={12},
  pages={12443--12451},
  year={2025}
}

@article{liu2026few,
  title={From few-shot to zero-shot: Towards generalist graph anomaly detection},
  author={Liu, Yixin and Li, Shiyuan and Zheng, Yu and Chen, Qingfeng and Zhang, Chengqi and Yu, Philip S and Pan, Shirui},
  journal={IEEE Transactions on Knowledge and Data Engineering},
  year={2026},
  publisher={IEEE}
}

@inproceedings{pan2025survey,
  title={A Survey of Generalization of Graph Anomaly Detection: From Transfer Learning to Foundation Models},
  author={Pan, Junjun and Zheng, Yu and Tan, Yue and Liu, Yixin},
  booktitle={The 16th IEEE International Conference on Knowledge Graphs},
  year={2025}
}

@inproceedings{miao2026blindguard,
  title={Blindguard: Safeguarding llm-based multi-agent systems under unknown attacks},
  author={Miao, Rui and Liu, Yixin and Wang, Yili and Shen, Xu and Tan, Yue and Dai, Yiwei and Pan, Shirui and Wang, Xin},
  booktitle={Proceedings of the 64th Annual Meeting of the Association for Computational Linguistics},
  year={2026}
}

@article{qian2026dynhd,
  title={DynHD: Hallucination Detection for Diffusion Large Language Models via Denoising Dynamics Deviation Learning},
  author={Qian, Yanyu and Tan, Yue and Liu, Yixin and Yu, Wang and Pan, Shirui},
  journal={arXiv preprint arXiv:2603.16459},
  year={2026}
}

@article{tan2024influence,
  title={Influence-oriented personalized federated learning},
  author={Tan, Yue and Long, Guodong and Jiang, Jing and Zhang, Chengqi},
  journal={arXiv preprint arXiv:2410.03315},
  year={2024}
}

@article{chen2025uncertainty,
  title={Uncertainty-aware graph neural networks: A multihop evidence fusion approach},
  author={Chen, Qingfeng and Li, Shiyuan and Liu, Yixin and Pan, Shirui and Webb, Geoffrey I and Zhang, Shichao},
  journal={IEEE Transactions on Neural Networks and Learning Systems},
  year={2025},
  publisher={IEEE}
}

@inproceedings{pan2026correcting,
  title={Correcting False Alarms from Unseen: Adapting Graph Anomaly Detectors at Test Time},
  author={Pan, Junjun and Liu, Yixin and Zhou, Chuan and Xiong, Fei and Liew, Alan Wee-Chung and Pan, Shirui},
  booktitle={Proceedings of the AAAI Conference on Artificial Intelligence},
  year={2026}
}

@article{yan2025address,
  title={Address Anomalies at Critical Crossroads for Graph Anomaly Detection},
  author={Yan, Junyi and Zuo, Enguang and Liang, Ke and Liu, Meng and Li, Miaomiao and Liu, Xinwang and Lv, Xiaoyi and Lu, Kai},
  journal={IEEE Transactions on Knowledge and Data Engineering},
  year={2025},
  publisher={IEEE}
}

@article{zuo2023sucola,
  title={SUCOLA: Self-adaptive structure refinement unsupervised contrastive learning framework for food safety risk early warning},
  author={Zuo, Enguang and Yan, Junyi and Aysa, Alimjan and Chen, Chen and Chen, Cheng and Ma, Hongbing and Lv, Xiaoyi and Ubul, Kurban},
  journal={Engineering Applications of Artificial Intelligence},
  volume={126},
  pages={107016},
  year={2023},
  publisher={Elsevier}
}

@article{zuo2025rethinking,
  title={Rethinking unsupervised time series anomaly detection: Dynamic attention based on route inverse-masking},
  author={Zuo, Enguang and Zhong, Jie and Chen, Chen and Chen, Cheng and Ubul, Kurban and Lv, Xiaoyi},
  journal={Applied Soft Computing},
  pages={113971},
  year={2025},
  publisher={Elsevier}
}

@inproceedings{pan2026camera,
  title={CAMERA: Adapting to Semantic Camouflage in Unsupervised Text-Attributed Graph Fraud Detection},
  author={Pan, Junjun and Liu, Yixin and Zheng, Yu and Chi, Lianhua and Liew, Alan Wee-Chung and Pan, Shirui},
  booktitle={International Joint Conference on Artificial Intelligence},
  year={2026}
}

@inproceedings{liu2026rethinking,
  title={Rethinking Feature Alignment in Generalist Graph Anomaly Detection: A Relational Fingerprint-based Approach},
  author={Liu, Yujing and Liu, Yixin and Zheng, Yu and Liew, Alan Wee-Chung and Cao, Xiaofeng and Pan, Shirui},
  booktitle={International Conference on Machine Learning},
  year={2026}
}

@inproceedings{zhao2026fedcigar,
  title={FedCIGAR: A Personalized Reconstruction Approach for Federated Graph-level Anomaly Detection},
  author={Zhao, Yunfeng and Liu, Yixin and Chen, Qingfeng and Li, Shiyuan and Tan, Yue and Pan, Shirui},
  booktitle={International Joint Conference on Artificial Intelligence},
  year={2026}
}

@article{shen2026raising,
  title={Raising the bar in graph ood generalization: Invariant learning beyond explicit environment modeling},
  author={Shen, Xu and Liu, Yixin and Wang, Yili and Miao, Rui and Dai, Yiwei and Pan, Shirui and Chang, Yi and Wang, Xin},
  journal={IEEE Transactions on Pattern Analysis and Machine Intelligence},
  year={2026}
}
